\pdfoutput=1

\documentclass[11pt]{article}

\usepackage[]{ACL2023}

\usepackage{times}
\usepackage{latexsym}
\usepackage{multirow}
\usepackage{array}
\usepackage{graphicx}
\usepackage{amssymb}
\usepackage{hhline}
\usepackage{amsmath}
\usepackage{cases}
\usepackage{mathtools}
\usepackage{enumitem}
\usepackage{diagbox}
\usepackage{booktabs}
\usepackage{subfigure}
\usepackage{url}
\usepackage{colortbl} 

\usepackage[T1]{fontenc}

\usepackage[utf8]{inputenc}

\usepackage{microtype}

\newcommand{\tabincell}[2]{\begin{tabular}{@{}#1@{}}#2\end{tabular}}
\definecolor{ggray}{HTML}{E7E6E6}
\definecolor{ggreen}{HTML}{d4e7cf}

\title{Towards Unifying Multi-Lingual and Cross-Lingual Summarization}

\author{Jiaan Wang\textsuperscript{1}\thanks{ \ \ Work was done when Jiaan Wang was interning at Pattern Recognition Center, WeChat AI, Tencent Inc, China.}, \ Fandong Meng\textsuperscript{2}, \ Duo Zheng\textsuperscript{3}, \ Yunlong Liang\textsuperscript{2}\\
\bf {Zhixu Li\textsuperscript{4}\thanks{ \ \ Corresponding authors.}, \ Jianfeng Qu\textsuperscript{1}\footnotemark[2] and \ Jie Zhou\textsuperscript{2}} \\
\small{\textsuperscript{1}School of Computer Science and Technology, Soochow University, Suzhou, China} \\
\small{\textsuperscript{2}Pattern Recognition Center, WeChat AI, Tencent Inc, China} \quad \small{\textsuperscript{3}Beijing University of Posts and Telecommunications} \\
\small{\textsuperscript{4}Shanghai Key Laboratory of Data Science, School of Computer Science, Fudan University, Shanghai, China} \\
\small \texttt{jawang.nlp@gmail.com}, \texttt{fandongmeng@tencent.com} \\
\small \texttt{zhixuli@fudan.edu.cn}, \texttt{jfqu@suda.edu.cn}
}

\begin{document}
\maketitle
\begin{abstract}
To adapt text summarization to the multilingual world, previous work proposes multi-lingual summarization (MLS) and cross-lingual summarization (CLS). However, these two tasks have been studied separately due to the different definitions, which limits the compatible and systematic research on both of them.
In this paper, we aim to unify MLS and CLS into a more general setting, \emph{i.e.}, many-to-many summarization (M2MS), where a single model could process documents in any language and generate their summaries also in any language.
As the first step towards M2MS, we conduct preliminary studies to show that M2MS can better transfer task knowledge across different languages than MLS and CLS.
Furthermore, we propose \textsc{Pisces}, a pre-trained M2MS model that learns language modeling, cross-lingual ability and summarization ability via three-stage pre-training.
Experimental results indicate that our \textsc{Pisces} significantly outperforms the state-of-the-art baselines, especially in the zero-shot directions, where there is no training data from the source-language documents to the target-language summaries.\footnote{\url{https://hf.co/Krystalan/PISCES}}

\end{abstract}

\section{Introduction}

\begin{figure}[t]
\centerline{\includegraphics[width=0.45\textwidth]{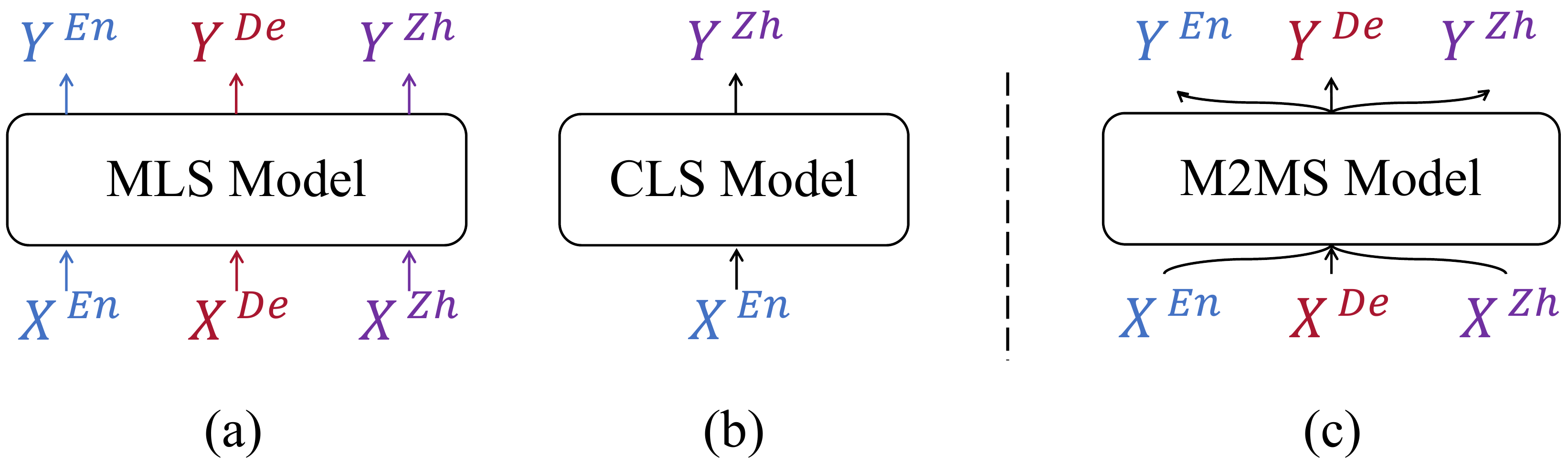}}
\caption{Illustration of (a) multi-lingual summarization, (b) cross-lingual summarization and (c) many-to-many summarization. $X^{i}$ and $Y^{i}$ denote the input document and output summary in language $i$, respectively. \textcolor[RGB]{60,103,188}{En}: English; \textcolor[RGB]{159,23,42}{De}: German; \textcolor[RGB]{101,42,150}{Zh}: Chinese.}
\label{fig:example}
\vspace{-0.3cm}
\end{figure}

The world we live in is multi-lingual. With globalization, text resources in various languages flood the Internet, where global users can easily access their desired information.
Under this background, the text summarization community presents multi-lingual summarization (MLS) and cross-lingual summarization (CLS), respectively. As shown in Figure~\ref{fig:example}, MLS aims at building a unified model to process documents in multiple languages and generate summaries in the corresponding language~\cite{giannakopoulos-etal-2015-multiling,Cao_Wan_Yao_Yu_2020,hasan-etal-2021-xl,wang-etal-2021-contrastive,varab-schluter-2021-massivesumm}, while CLS generates a summary in the target language from the given document in a different source language~\cite{10.1145/979872.979877,wan-etal-2010-cross,wan-2011-using,yao-etal-2015-phrase,zhu-etal-2019-ncls,ladhak-etal-2020-wikilingua,perez-beltrachini-lapata-2021-models,Wang2022ClidSumAB,Wang2022ASO,wang2022understanding,Wang2023ZeroShotCS}.
Despite the close relationship between MLS and CLS (\emph{e.g.}, both tasks involve more than one language and require models to distill the key information from documents), previous work studies each task separately, hindering the systematic exploration for both of them.

In this paper, we aim to unify MLS and CLS into a more general setting named \emph{many-to-many summarization} (M2MS).
As its name implies, the goal of M2MS is to build a single summarization model to process a document in any source language and generate the corresponding summary in any given target language.
In this manner, one M2MS model could perform more directions than MLS and CLS\footnote{We use ``direction'' to denote the summarization direction from the source to the target languages, e.g., English (documents) $\Rightarrow$ Chinese (summaries).}, thus reducing the used parameters. For example, one M2MS model involving $n$ languages could replace one MLS model and $n$$\times$$(n-1)$ CLS models.
To provide a deeper understanding of M2MS, we also conduct preliminary studies to systematically compare M2MS with MLS and CLS, respectively.
In detail, following recent CLS work~\cite{ladhak-etal-2020-wikilingua,perez-beltrachini-lapata-2021-models}, we use mBART-50~\cite{Tang2020MultilingualTW} as the summarization model, and train the model in the settings of MLS, CLS and M2MS, respectively.
After comparing the model performances, we find that the model trained in M2MS setting can better transfer task knowledge across different languages and combine the advantages of those trained in MLS and CLS settings.
Therefore, we argue that it is promising to unify MLS and CLS into M2MS.

Furthermore, we propose \textsc{Pisces}\footnote{\textsc{Pisces}: \textbf{P}re-tra\textbf{I}ning with gap-\textbf{S}entences and \textbf{C}ross-lingual d\textbf{E}noi\textbf{S}ing for many-to-many summarization.}, a pre-trained M2MS model that learns language modeling, cross-lingual ability and summarization ability via three pre-training stages:
(1) \textit{meta pre-training} learns the general language modeling knowledge from multi-lingual unlabeled corpora; (2) \textit{cross-lingual pre-training} makes the model aware of the transformation between different languages based on parallel corpora; (3) \textit{task-specific pre-training} utilizes M2MS objective to simultaneously improve the cross-lingual ability and the summarization abilities of the model.
Considering the high-quality M2MS samples are non-trivial to collect, we leverage a simple strategy to construct pseudo M2MS samples from multi-lingual unlabeled corpora.
During the three-stage pre-training, \textsc{Pisces} gradually shifts from learning language modeling to the abilities required by M2MS.
Among them, the learned cross-lingual ability plays a key role in enhancing the knowledge transferability of the downstream task (\emph{i.e.}, summarization) from high-resource languages to low/zero-resource languages.
Lastly, the pre-trained \textsc{Pisces} could be simply fine-tuned on M2MS with input source-language documents and output target-language summaries.

We evaluate \textsc{Pisces} on the WikiLingua~\cite{ladhak-etal-2020-wikilingua} and CrossSum~\cite{Hasan2021CrossSumBE} datasets.
Experimental results show that \textsc{Pisces} achieves promising results compared with the state-of-the-art baselines (\emph{i.e.}, mBART-50 and mT5), especially in the zero-shot directions.
Moreover, we find that \textsc{Pisces} is even able to generate summaries for documents whose language never occurs in the fine-tuning stage.

Our contributions are concluded as follows:
\begin{itemize}[leftmargin=*,topsep=0pt]
\setlength{\itemsep}{0pt}
\setlength{\parsep}{0pt}
\setlength{\parskip}{0pt}
    \item To our knowledge, we are the first to unify MLS and CLS into a more general setting (M2MS). We also conduct preliminary studies to provide deeper analyses among MLS, CLS and M2MS.
    \item We propose \textsc{Pisces}, a pre-trained M2MS model that learns language modeling, cross-lingual ability and summarization ability through a carefully designed three-stage pre-training.
    \item We conduct extensive experiments and show that our \textsc{Pisces} achieves new state-of-the-art performance on the large-scale benchmark datasets. Besides, the effectiveness of \textsc{Pisces} in low/zero-resource languages is also demonstrated.
\end{itemize}

\begin{table*}[t]
\centering
\resizebox{0.98\textwidth}{!}
{
\begin{tabular}{c|c|c|c|c|c|c|c}
\hline
                \diagbox[dir=NW]{Src}{Trg} & Setting & En                                                     & Fr                                                      & Hi                                                     & Zh                                                      & Th                                                      & Tr                                                    \\ \hline
         & \texttt{ONE}              & 41.2 / 17.5 / 34.6 / 74.2                        & 35.2 / 14.8 / 29.2 / 73.0                         & 28.2 / \textcolor{white}{0}8.3 / 22.6 / 67.7                         & 34.9 / 11.8 / 30.4 / 69.8                        & 34.3 / 14.3 / 30.0 / 66.1                      & \cellcolor{ggray!90}NA             \\ 
                     & \texttt{U-CLS}             & 39.7 / 16.0 / 32.7 / 73.6      & \underline{36.8} / \underline{15.3} / \underline{29.9} / \underline{73.6}                     & \underline{31.2} / \textcolor{white}{0}\underline{9.2} / \underline{23.9} / \underline{69.0}        & \textbf{37.9} / \textbf{13.9} / \textbf{32.7} / \textbf{71.5}                      & \underline{38.9} / \underline{17.9} / \underline{33.4} / \underline{68.9}                 & \cellcolor{ggray!90}\textbf{3.2} / \textbf{0.3} / \textbf{3.0} / \underline{48.9} \\
                     & \texttt{MLS}             & \underline{41.6} / \underline{17.9} / \underline{34.7} / \underline{74.4}                        & \textcolor{white}{0}5.3 / \textcolor{white}{0}0.8 / \textcolor{white}{0}4.8 / 63.8 & \textcolor{white}{0}3.3 / \textcolor{white}{0}0.7 / \textcolor{white}{0}3.1 / 53.7                         & 14.6 / \textcolor{white}{0}0.9 / 14.5 / 60.1                        & 20.8 / \textcolor{white}{0}5.7 / 20.0 / 54.1       & \cellcolor{ggray!90}2.5 / \underline{0.2} / 2.4 / 47.3 \\
                  \multirow{-4}{*}{En}   & \texttt{M2MS}             & \textbf{41.9} / \textbf{18.2} / \textbf{34.9} / \textbf{74.6}           & \textbf{37.2} / \textbf{15.8} / \textbf{30.3} / \textbf{73.9}    & \textbf{31.7} / \textcolor{white}{0}\textbf{9.6} / \textbf{24.5} / \textbf{69.3}     & \textbf{37.9} / \textbf{13.9} / \textbf{32.7} / \textbf{71.5}        & \textbf{39.5} / \textbf{18.5} / \textbf{34.0} / \textbf{69.1}     & \cellcolor{ggray!90}\textbf{3.2} / \underline{0.2} / \textbf{3.0} / \textbf{49.0}  \\ \hline
                    & \texttt{ONE}              & 35.6 / 13.6 / 29.8 / 72.1       & 37.8 / 17.4 / 31.2 / 73.9                        & \cellcolor{ggray!90}NA              & 32.6 / 10.0 / 28.4 / 68.6                        & 31.4 / 11.8 / 27.6 / 64.9                       & \cellcolor{ggray!90}NA             \\ 
                     & \texttt{U-CLS}             & \underline{37.5} / \underline{14.4} / \underline{30.7} / \underline{72.9}     & 37.6 / 16.1 / 30.5 / 74.0  & \cellcolor{ggray!90}\underline{28.2} / \textcolor{ggray!90}{0}\underline{7.6} / \underline{22.0} / \textbf{68.1} & \underline{36.7} / \textbf{12.8} / \underline{31.3} / \textbf{70.9}         & \underline{37.3} / \underline{16.2} / \underline{32.1} / \underline{68.1}                         & \cellcolor{ggray!90}\textbf{3.3} / \textbf{0.3} / \textbf{3.1} / \textbf{49.4} \\
                     & \texttt{MLS}            & \textcolor{white}{0}8.8 / \textcolor{white}{0}2.2 / \textcolor{white}{0}7.6 / 64.3          & \textbf{39.5} / \textbf{18.2} / \textbf{32.5} / \textbf{74.9}                         & \cellcolor{ggray!90}\textcolor{ggray!90}{0}2.1 / \textcolor{ggray!90}{0}0.4 / \textcolor{ggray!90}{0}1.9 / 53.3 & 13.5 / \textcolor{white}{0}1.0 / 13.2 / 57.5                      & 18.5 / \textcolor{white}{0}3.3 / 17.9 / 54.5                       & \cellcolor{ggray!90}2.1 / 0.1 / 2.1 / 46.8 \\
                    \multirow{-4}{*}{Fr} & \texttt{M2MS}             & \textbf{38.2} / \textbf{15.0} / \textbf{31.7} / \textbf{73.4}                    & \underline{39.2} / \underline{17.9} / \underline{32.0} / \underline{74.7}                     & \cellcolor{ggray!90}\textbf{28.7} / \textcolor{ggray!90}{0}\textbf{7.9} / \textbf{22.3} / \textbf{68.1} & \textbf{36.9} / \textbf{12.8} / \textbf{31.6} / \textbf{70.9}       &  \textbf{37.9} / \textbf{16.6} / \textbf{32.6} / \textbf{68.5}         & \cellcolor{ggray!90}\underline{3.1} / \underline{0.2} / \underline{3.0} / \underline{49.2} \\ \hline
                     & \texttt{ONE}              & 32.2 / 10.9 / 26.1 / 70.2            & \cellcolor{ggray!90}NA               & 32.8 / 11.5 / 25.8 / 69.6                       & \cellcolor{ggray!90}NA               & \cellcolor{ggray!90}NA               & \cellcolor{ggray!90}NA             \\ 
                     & \texttt{U-CLS}             & \underline{36.8} / \underline{14.0} / \underline{29.8} / \underline{72.2}       & \cellcolor{ggray!90}\underline{31.9} / \underline{11.6} / \underline{24.7} / \underline{71.4}  & 32.7 / 10.3 / 25.6 / 70.3  & \cellcolor{ggray!90}\underline{32.6} / \underline{10.2} / \underline{27.3} / \underline{68.6} & \cellcolor{ggray!90}\underline{34.9} / \underline{14.3} / \underline{29.4} / \underline{67.1} & \cellcolor{ggray!90}\underline{3.3} / \textbf{0.3} / \textbf{3.2} / \textbf{50.0} \\
                     & \texttt{MLS}             & 11.1 / \textcolor{white}{0}3.3 / \textcolor{white}{0}9.3 / 57.7                      & \cellcolor{ggray!90}11.6 / \textcolor{ggray!90}{0}3.2 / \textcolor{ggray!90}{0}9.5 / 59.3 & \textbf{36.0} / \textbf{12.7} / \textbf{27.8} / \textbf{71.3}          & \cellcolor{ggray!90}14.2 / \textcolor{ggray!90}{0}2.8 / 12.8 / 57.2 & \cellcolor{ggray!90}23.1 / \textcolor{ggray!90}{0}6.0 / 21.3 / 57.9 & \cellcolor{ggray!90}2.1 / 0.1 / 2.0 / 46.7 \\
                    \multirow{-4}{*}{Hi} & \texttt{M2MS}             & \textbf{37.9} / \textbf{14.6} / \textbf{30.8} / \textbf{72.8}           & \cellcolor{ggray!90}\textbf{32.8} / \textbf{12.2} / \textbf{25.9} / \textbf{72.1} & \underline{35.6} / \underline{12.5} / \textbf{27.8} / \underline{71.1} & \cellcolor{ggray!90}\textbf{33.2} / \textbf{10.6} / \textbf{28.2} / \textbf{69.1} & \cellcolor{ggray!90}\textbf{35.4} / \textbf{14.6} / \textbf{30.1} / \textbf{67.4} & \cellcolor{ggray!90}\textbf{3.4} / \textbf{0.3} / \textbf{3.2} / \underline{49.7} \\ \hline
                     & \texttt{ONE}              & 34.6 / 11.8 / 28.4 / 71.4                     & 31.5 / 11.4 / 25.4 / 71.0                       & \cellcolor{ggray!90}NA              & 40.8 / 16.9 / 35.4 / 71.9                        & \cellcolor{ggray!90}NA               & \cellcolor{ggray!90}NA             \\ 
                     & \texttt{U-CLS}             & \underline{37.7} / \underline{14.1} / \underline{30.8} / \underline{72.8}        & \underline{35.4} / \underline{14.1} / \underline{28.4} / \underline{73.0}   & \cellcolor{ggray!90}\underline{25.8} / \textcolor{ggray!90}{0}\underline{6.1} / \underline{20.0} / \underline{66.4}  & 39.6 / 15.1 / 34.2 / 72.2   & \cellcolor{ggray!90}\underline{36.6} / \textbf{15.3} / \underline{31.0} / \underline{67.3} & \cellcolor{ggray!90}\underline{3.3} / \textbf{0.2} / \underline{3.1} / \textbf{49.8}  \\
                     & \texttt{MLS}             & 10.4 / \textcolor{white}{0}3.0 / \textcolor{white}{0}8.6 / 61.7  & 24.9 / \textcolor{white}{0}7.3 / 19.7 / 68.0      & \cellcolor{ggray!90}20.4 / \textcolor{ggray!90}{0}4.4 / 16.0 / 62.4 & \textbf{42.8} / \textbf{17.9} / \textbf{37.0} / \textbf{73.1}    & \cellcolor{ggray!90}30.3 / \textcolor{ggray!90}{0}9.3 / 26.4 / 63.5  & \cellcolor{ggray!90}2.8 / \textbf{0.2} / 2.6 / 48.4 \\
                    \multirow{-4}{*}{Zh} & \texttt{M2MS}             & \textbf{39.2} / \textbf{15.1} / \textbf{32.0} / \textbf{73.4}                    & \textbf{36.0} / \textbf{14.5} / \textbf{29.0} / \textbf{73.3}     & \cellcolor{ggray!90}\textbf{27.0} / \textcolor{ggray!90}{0}\textbf{6.6} / \textbf{20.8} / \textbf{66.9}  &  \underline{41.7} / \underline{17.0} / \underline{35.9} / \underline{72.7}  & \cellcolor{ggray!90}\textbf{36.8} / \textbf{15.3} / \textbf{31.4} / \textbf{67.6}  & \cellcolor{ggray!90}\textbf{3.4} / \textbf{0.2} / \textbf{3.2} / \underline{49.6}  \\ \hline
                     & \texttt{ONE}              & 32.1 / 11.1 / 26.4 / 70.4          & 27.9 / \textcolor{white}{0}2.7 / 22.7 / 69.4        & \cellcolor{ggray!90}NA              & \cellcolor{ggray!90}NA               & 37.8 / 17.6 / 33.0 / 67.4                        & \cellcolor{ggray!90}NA             \\ 
                     & \texttt{U-CLS}             & \underline{37.2} / \underline{14.4} / \underline{30.7} / \underline{72.6}            & \underline{34.9} / \underline{13.9} / \underline{27.7} / \underline{72.3}                      & \cellcolor{ggray!90}\underline{27.1} / \textcolor{ggray!90}{0}\underline{6.8} / \underline{20.6} / \underline{66.9}  & \cellcolor{ggray!90}\underline{34.1} / \underline{10.9} / \underline{28.3} / \underline{68.9}  & 39.9 / 18.4 / 34.3 / 69.5            & \cellcolor{ggray!90}\textbf{3.4} / \textbf{0.3} / \textbf{3.2} / \textbf{49.4} \\
                     & \texttt{MLS}            & \textcolor{white}{0}7.4 / \textcolor{white}{0}1.8 / \textcolor{white}{0}6.6 / 54.9        & 10.1 / \textcolor{white}{0}2.5 / \textcolor{white}{0}8.4 / 58.4                         & \cellcolor{ggray!90}11.8 / \textcolor{ggray!90}{0}2.1 / \textcolor{ggray!90}{0}9.6 / 57.6 & \cellcolor{ggray!90}16.8 / \textcolor{ggray!90}{0}3.3 / 15.0 / 59.4 & \textbf{43.3} / \textbf{22.3} / \textbf{37.1} / \textbf{70.3}   & \cellcolor{ggray!90}2.7 / \textbf{0.3} / 2.6 / 47.8 \\
                    \multirow{-4}{*}{Th} & \texttt{M2MS}             & \textbf{38.5} / \textbf{15.4} / \textbf{31.9} / \textbf{73.4}       & \textbf{35.6} / \textbf{14.2} / \textbf{28.3} / \textbf{72.9}    & \cellcolor{ggray!90}\textbf{27.8} / \textcolor{ggray!90}{0}\textbf{7.3} / \textbf{21.4} / \textbf{67.4}  & \cellcolor{ggray!90}\textbf{34.6} / \textbf{11.3} / \textbf{29.0} / \textbf{69.4}  & \underline{42.2} / \underline{20.8} / \underline{36.2} / \underline{70.1}  & \cellcolor{ggray!90}\underline{3.3} / \textbf{0.3} / \underline{3.1} / \underline{49.3}  \\ \hline
                    & \texttt{ONE}              & \cellcolor{ggray!90}NA              & \cellcolor{ggray!90}NA               & \cellcolor{ggray!90}NA              & \cellcolor{ggray!90}NA               & \cellcolor{ggray!90}NA               & \cellcolor{ggray!90}NA            \\ 
                     & \texttt{U-CLS}             & \cellcolor{ggray!90}\textbf{16.9} / \textcolor{ggray!90}{0}\textbf{3.3} / \textbf{14.4} / \textbf{62.9}   & \cellcolor{ggray!90}\textbf{16.7} / \textcolor{ggray!90}{0}\textbf{3.3} / \textbf{13.5} / \textbf{64.6}   & \cellcolor{ggray!90}\textbf{16.2} / \textcolor{ggray!90}{0}\textbf{2.6} / \textbf{13.7} / \textbf{61.0}   & \cellcolor{ggray!90}\textbf{21.7} / \textcolor{ggray!90}{0}\textbf{3.8} / \textbf{19.1} / \textbf{61.2}   & \cellcolor{ggray!90}\textbf{22.8} / \textcolor{ggray!90}{0}\textbf{5.7} / \textbf{19.9} / \textbf{60.4}   & \cellcolor{ggray!90}\textbf{3.4} / \textbf{0.3} / \textbf{3.3} / \textbf{48.8}  \\
                     & \texttt{MLS}             & \cellcolor{ggray!90}\textcolor{ggray!90}{0}6.6 / \textcolor{ggray!90}{0}0.8 / \textcolor{ggray!90}{0}5.9 / 53.5  & \cellcolor{ggray!90}\textcolor{ggray!90}{0}9.7 / \textcolor{ggray!90}{0}1.1 / \textcolor{ggray!90}{0}8.6 / 58.7    & \cellcolor{ggray!90}\textcolor{ggray!90}{0}7.8 / \textcolor{ggray!90}{0}0.7 / \textcolor{ggray!90}{0}7.0 / 54.1  & \cellcolor{ggray!90}17.9 / \textcolor{ggray!90}{0}2.8 / 15.3 / 58.7  & \cellcolor{ggray!90}17.4 / \textcolor{ggray!90}{0}2.5 / 16.6 / 54.4  & \cellcolor{ggray!90}2.3 / 0.1 / 2.2 / 44.7  \\
                    \multirow{-4}{*}{Tr}  & \texttt{M2MS}            & \cellcolor{ggray!90}\underline{15.7} / \textcolor{ggray!90}{0}\underline{2.6} / \underline{13.4} / \underline{62.1}  & \cellcolor{ggray!90}\underline{16.0} / \textcolor{ggray!90}{0}\underline{3.2} / \underline{13.2} / \underline{64.4}  & \cellcolor{ggray!90}\underline{14.9} / \textcolor{ggray!90}{0}\underline{2.3} / \underline{12.6} / \underline{60.1}  & \cellcolor{ggray!90}\underline{19.9} / \textcolor{ggray!90}{0}\underline{3.0} / \underline{17.6} / \underline{60.0}  & \cellcolor{ggray!90}\underline{21.4} / \textcolor{ggray!90}{0}\underline{4.8} /  \underline{19.3} / \underline{59.9}  & \cellcolor{ggray!90}\underline{3.1} / \underline{0.2} / \underline{3.0} / \underline{48.4}   \\ \hline
\end{tabular}

}

\caption{Results on WikiLingua (\textsc{Rouge-1} / \textsc{Rouge-2} / \textsc{Rouge-l} / \textsc{BertScore}). Since there is no training data in \colorbox{ggray!90}{zero-shot} directions, mBART (\texttt{ONE}) cannot be trained and we denote the results as ``NA''. The \textbf{bold} and \underline{underline} denote the best and the second-best scores, respectively.} 
\label{table:overview_e2e}
\vspace{-0.3cm}
\end{table*}

\section{Related Work}

\noindent \textbf{Multi-Lingual Summarization.}
Multi-lingual summarization (MLS) aims to process documents in multiple languages and generate their summaries in the corresponding language. \citet{giannakopoulos-etal-2015-multiling} present MultiLing-2015 dataset. Later, this task receives increasing attention~\cite{Vanetik2015MultilingualSW,litvak-etal-2016-museec}. Recently, large-scale MLS datasets~\cite{scialom-etal-2020-mlsum,varab-schluter-2021-massivesumm,hasan-etal-2021-xl,feng-etal-2022-msamsum,liang2022summary} together with sophisticated methods~\cite{Cao_Wan_Yao_Yu_2020,Chi2020CrossLingualNL,wang-etal-2021-contrastive} are proposed one after another.
Considering the close relation between MLS and CLS, \citet{Cao_Wan_Yao_Yu_2020,feng-etal-2022-msamsum} also evaluate the MLS models on CLS to show their zero-shot CLS ability.

\vspace{0.5ex}
\noindent \textbf{Cross-Lingual Summarization.}
Given documents in one language, cross-lingual summarization (CLS) generates summaries in another language.
Early work typically focuses on pipeline methods~\cite{Leuski2003CrosslingualCE,Orasan2008EvaluationOA,wan-etal-2010-cross,wan-2011-using,yao-etal-2015-phrase}, \emph{i.e.}, translation and then summarization or summarization and then translation.
Recently, with the availability of large-scale CLS datasets~\cite{zhu-etal-2019-ncls,ladhak-etal-2020-wikilingua,perez-beltrachini-lapata-2021-models,Wang2022ClidSumAB,chen2022cross,10.1145/3539597.3570479},
many researchers shift the research attention to end-to-end CLS models, including multi-task learning~\cite{cao-etal-2020-jointly,Bai2021BridgingTG,Liang2022AVH}, knowledge distillation~\cite{Nguyen2021ImprovingNC}, resource-enhanced~\cite{zhu-etal-2020-attend} and pre-training~\cite{xu-etal-2020-mixed,chi-etal-2021-mt6} approaches.
Among them, most CLS work separately builds CLS models in each cross-lingual direction except for \citet{Hasan2021CrossSumBE}, who jointly train mT5~\cite{xue-etal-2021-mt5} in multiple directions.

Different from previous MLS and CLS, we unify them into a more general setting (M2MS) starting from the training stage. Besides, we are the first to systematically investigate the capabilities of models trained with MLS, CLS and M2MS settings.

\vspace{0.5ex}
\noindent \textbf{Pre-Trained Models for Summarization.}
Pre-trained models have shown their superiority in summarization task, \emph{e.g.}, BART~\cite{lewis-etal-2020-bart} and T5~\cite{JMLR:v21:20-074}.
To enhance the summarization ability during the pre-training stage, \textsc{Pegasus}~\cite{pmlr-v119-zhang20ae} introduces the gap sentence generation (GSG) objective to enable the model to generate key sentences in an article from the remaining ones.
Further, \textsc{Primera}~\cite{Xiao2021PRIMERPM} extends GSG from single-document to multi-document summarization.
In dialogue scenarios, \citet{Wang2022ClidSumAB} present m\textsc{Dial}BART for cross-lingual dialogue summarization.

Among these pre-trained summarization models, \textsc{Pegasus} and \textsc{Primera} only focus on monolingual summarization. Though m\textsc{Dial}BART aims at CLS, the model is merely built for a single cross-lingual direction (\emph{i.e.}, English $\Rightarrow$ German/Chinese) and a specific scenario (\emph{i.e.}, dialogue).
Our \textsc{Pisces} is the first multi-lingual pre-trained model for general summarization.

\section{Does Unifying All Directions in a Single Model Help Each Other?}
\label{sec:3}

As discussed previously, M2MS unifies all summarization directions in a single model.
Therefore, we wonder \textit{can such a setting help the model better transfer task knowledge across different languages compared with the settings of MLS and CLS?}
To answer the question, we conduct preliminary studies to investigate the influence of different settings.

\subsection{Setup}

\noindent \textbf{Data.} The preliminary studies are conducted on WikiLingua~\cite{ladhak-etal-2020-wikilingua}, one of the largest CLS datasets.
We focus on six languages, \emph{i.e.}, English (En), French (Fr), Hindi (Hi), Chinese (Zh), Thai (Th) and Turkish (Tr).
Among them, Tr serves as a zero-resource language, whose documents and summaries only appear in the validation and test sets.
More details are given in Section~\ref{subsec:datasets_and_metrics}.

\vspace{0.5ex}
\noindent \textbf{Summarization Model.} Following recent CLS literature~\cite{ladhak-etal-2020-wikilingua,perez-beltrachini-lapata-2021-models}, we use mBART-50~\cite{Tang2020MultilingualTW} as the summarization model, and train the model in the following four settings:
\begin{itemize}[leftmargin=*,topsep=0pt]
\setlength{\itemsep}{0pt}
\setlength{\parsep}{0pt}
\setlength{\parskip}{0pt}
\item mBART (\texttt{ONE}): We separately train several models, each of which is built and evaluated in one single direction. When the direction is cross-lingual (or monolingual), the corresonding model is a CLS (or monolingual summarization) model.
\item mBART (\texttt{U-CLS}): We train a unified model with all cross-lingual samples, and test the model in all directions.
\item mBART (\texttt{MLS}): We train one unified model with monolingual samples in all languages. Then, the trained model is evaluated in all directions.
\item mBART (\texttt{M2MS}): It is a new setting introduced by this work, where the model is both trained and evaluated in all directions. 
\end{itemize}

\subsection{Analytic Results}
Table~\ref{table:overview_e2e} shows the results in terms of \textsc{Rouge}~\cite{Lin2004ROUGEAP} and \textsc{BertScore}~\cite{Zhang2020BERTScoreET}.

\vspace{0.5ex}
\noindent \textbf{mBART} (\texttt{M2MS}) \textbf{vs. mBART} (\texttt{CLS})\textbf{.}
The results in all directions show that mBART (\texttt{M2MS}) outperforms mBART (\texttt{CLS}) in all metrics, illustrating that unifying all directions in a single model could transfer task knowledge across different languages.

\vspace{0.5ex}
\noindent \textbf{mBART} (\texttt{M2MS}) \textbf{vs. mBART} (\texttt{MLS})\textbf{.} Comparing mBART (\texttt{M2MS}) and mBART (\texttt{MLS}), it is apparent to find that mBART (\texttt{M2MS}) significantly outperforms mBART (\texttt{MLS}) in cross-lingual directions (\emph{e.g.}, 26.9 vs. 11.7 \textsc{Rouge-1} in average), while achieving competitive results in monolingual directions (\emph{e.g.}, 33.9 vs. 34.2 \textsc{Rouge-1} in average).

To give a deeper understanding of why mBART (\texttt{MLS}) performs poorly in cross-lingual directions, we analyze its generated summaries and find that most of them are not in the language we expected.
Table~\ref{table:off_target} shows the rate of the generated summaries in the correct language.\footnote{Other directions also show similar situations.}
The languages of the generated summaries are detected by \textit{fastlangid}\footnote{\url{https://pypi.org/project/fastlangid/}}.
Compared with mBART (\texttt{M2MS}), mBART (\texttt{MLS}) struggles to generate summaries in the target language.
We conjecture this is because that mBART (\texttt{MLS}) is only trained with monolingual data from multiple languages without any cross-lingual signals, resulting in limited cross-lingual ability.

Based on the above analyses, we argue that the summarization signals from cross-lingual directions could help mBART (\texttt{M2MS}) perform CLS and transfer the task knowledge to zero-shot directions, while mBART (\texttt{MLS}) does not own such abilities.

\begin{table}[t]
\centering
\resizebox{0.48\textwidth}{!}
{
\begin{tabular}{l|cccc}
\hline
      & En$\Rightarrow$Fr & En$\Rightarrow$Hi & En$\Rightarrow$Zh & En$\Rightarrow$Th \\
mBART (\texttt{MLS}) & 5.8   & 0.2   & 1.3   & 1.0   \\
mBART (\texttt{M2MS}) & 99.9  & 99.4  & 95.4  & 99.9  \\ \hline
      & Fr$\Rightarrow$Hi & Fr$\Rightarrow$Zh & Fr$\Rightarrow$Th & Th$\Rightarrow$En \\
mBART (\texttt{MLS}) & 5.3   & 5.6   & 9.4   & 8.2   \\
mBART (\texttt{M2MS}) & 99.4  & 95.8  & 99.9  & 99.5  \\ \hline
\end{tabular}

}
\caption{Correct language rate (\%) of the summaries generated by mBART (\texttt{MLS}) and mBART (\texttt{M2MS}).}
\label{table:off_target}
\vspace{-0.3cm}
\end{table}

\vspace{0.5ex}
\noindent \textbf{mBART} (\texttt{M2MS}) \textbf{vs. mBART} (\texttt{U-CLS})\textbf{.}
The only difference between mBART (\texttt{M2MS}) and mBART (\texttt{U-CLS}) is that the training data of mBART (\texttt{M2MS}) contains all monolingual samples, while mBART (\texttt{U-CLS}) does not.
We find that the performance gap between mBART (\texttt{M2MS}) and mBART (\texttt{U-CLS}) is extremely smaller than that between mBART (\texttt{M2MS}) and mBART (\texttt{CLS}) / mBART (\texttt{MLS}).
In detail, mBART (\texttt{M2MS}) outperforms mBART (\texttt{U-CLS}) in most directions when the source and the target languages have been seen during the fine-tuning stage, \emph{i.e.}, the source and the target languages are from $\{$En, Fr, Hi, Zh, Th$\}$.
However, when the source or target language is unseen (\emph{i.e.}, Tr), the performance of mBART (\texttt{M2MS}) is slightly worse than mBART (\texttt{CLS}). This is because the monolingual training data used in mBART (\texttt{M2MS}) makes the word embeddings of the unseen language\footnote{We use ``unseen language'' to indicate the language does not occur in the \emph{fine-tuning} stage.} drift away from those of other languages (see details in Appendix~\ref{appendix:token_representation}). Additionally, the cross-lingual signal between the unseen language and other languages never occurs in the fine-tuning stage, making it difficult to summarize from or to the unseen language.

\subsection{Preliminary Conclusion}
The preliminary studies comparing mBART trained in different settings indicate that (1) the multi-lingual model trained in M2MS setting can better transfer task knowledge across different languages than those trained in the settings of MLS, CLS and unified CLS.
(2) Compared with unified CLS, M2MS helps the model achieve better transferability across visible languages, but sacrifices the transferability to unseen languages.

Grounding the above analyses, we argue that it is valuable to unify previous MLS and CLS to M2MS.
Meanwhile, \textit{how to improve the transferability to unseen languages} becomes a keypoint in M2MS.

\begin{figure*}[t]
\centerline{\includegraphics[width=0.98\textwidth]{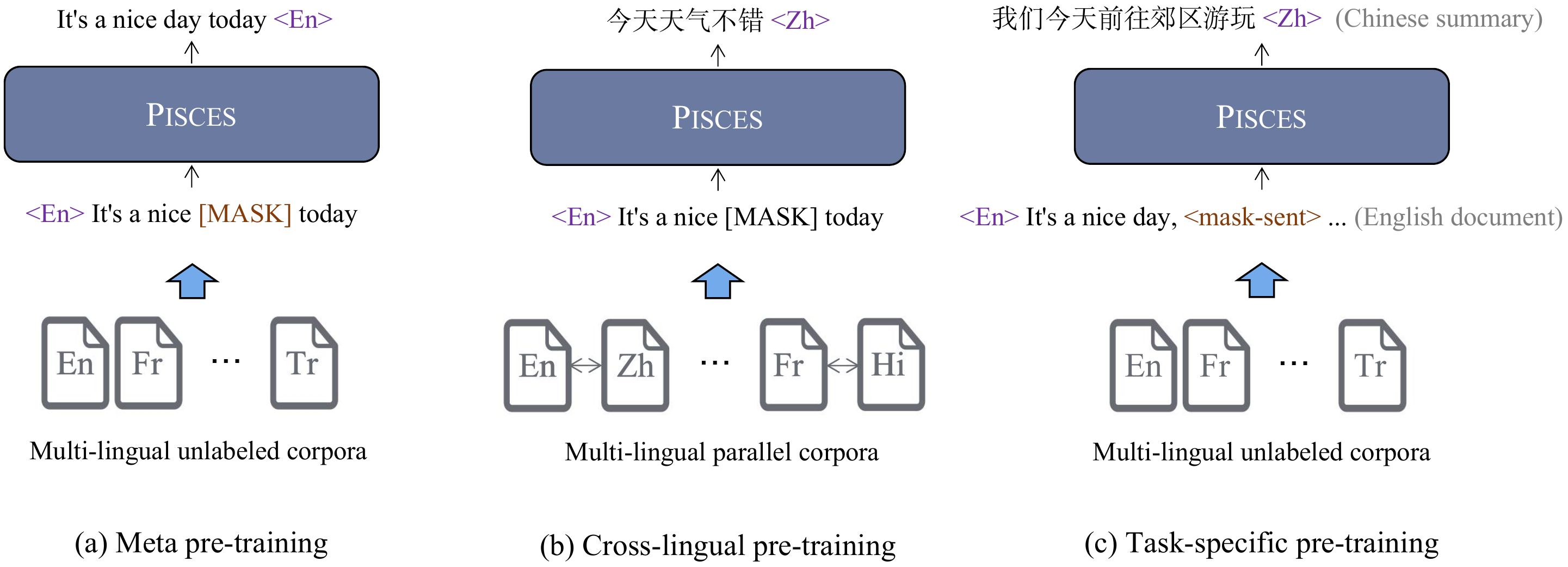}}
\caption{Overview of the three-stage pre-training in \textsc{Pisces}. Specifically, (a) meta pre-training requires the model to generate original sentences based on the noisy counterparts; (b) cross-lingual pre-training generates the sentences in the target language based on the noisy parallel sentences in the source language; (c) task-specific pre-training utilizes pseudo M2MS samples to pre-train the model.}
\label{fig:pisces}
\vspace{-0.3cm}
\end{figure*}

\section{\textsc{Pisces}}
In this section, we propose \textsc{Pisces}, a pre-trained multi-lingual model for M2MS with the backbone of transformer~\cite{vaswani2017attention}.

Figure~\ref{fig:pisces} shows the overview of \textsc{Pisces}, which contains three pre-training stages. Specifically, the meta pre-training (\S~\ref{subsec:meta_pretraining}) lets the pre-trained model learn general language modeling via monolingual denoising objective in multiple languages.
Then, to improve the transferability across different languages, the cross-lingual pre-training (\S~\ref{subsec:cross_lingual_pretraining}) adds noises to the source-language sentences, and encourages the model to translate them into parallel sentences in the target language.
Note that the parallel sentences used in this stage might involve the languages which are not seen in downstream tasks, and it is the key to improving the transferability to these languages.
Finally, to narrow the gap between the pre-training and fine-tuning stages, the task-specific pre-training (\S~\ref{subsec:task_spe_pretraining}) trains the model with pseudo M2MS samples, which are constructed from the multi-lingual unlabeled corpora via gap sentences selection and machine translation.
During the three-stage pre-training process, the model gradually learns the ability of language modeling, then the cross-lingual ability, and finally the adaptation to the specific task.

\subsection{Meta Pre-Training}
\label{subsec:meta_pretraining}
The goal of meta pre-training is to provide good initialization for the subsequent pre-training stages.
Here, we directly utilize mBART-50~\cite{Tang2020MultilingualTW} as the meta pre-trained model.

mBART-50 is a multi-lingual BART~\cite{lewis-etal-2020-bart} with the transformer encoder-decoder architecture.
The model is pre-trained on large-scale multi-lingual unlabeled corpora to learn the multi-lingual language modeling.
Specifically, following BART, the denoising task is used as the pre-training objective, and there are two types of noise: (1) \textit{text infilling} randomly masks text spans in text sequences, and (2) \textit{sentence permutation} randomly shuffles sentences in documents.
The model is required to comprehend the noisy text sequences and recover them.
To indicate the input and output languages, the language tags (\emph{e.g.}, \texttt{<En>} and \texttt{<Zh>}) are appended at the inputs of encoder and decoder sides, respectively.

\subsection{Cross-Lingual Pre-Training}
\label{subsec:cross_lingual_pretraining}
Despite the effectiveness of mBART-50, the input and output sequences in its pre-training stage are always in the same language, resulting in the under-explored cross-lingual ability. However, such ability is indispensable for M2MS.
Therefore, cross-lingual pre-training is designed to improve the cross-lingual transferability.

In detail, we propose a simple yet effective pre-training task, \emph{i.e.}, cross-lingual denoising, which lets the model generate sentences in the target language based on their noisy parallel sentences in a different source language.
The noise used in this stage is \textit{text infilling}.
In this way, the pre-trained model is required to not only understand the text in the source language but also learn the transformation between different languages.

\begin{table*}[t]
\centering
\resizebox{0.98\textwidth}{!}
{
\begin{tabular}{clcccccc}
\midrule[1pt]
\diagbox[dir=NW]{Src}{Trg}                    &                & En                    & Fr                    & Hi        &  Zh & Th & Tr            \\ \bottomrule[1pt]
\multirow{2}{*}{En} & \# Samples   & \cellcolor{ggreen!90}{124589 / 8351 / 8517}  & \cellcolor{ggreen!90}{53232 / 5161 / 5258}   & \cellcolor{ggreen!30}{5707 / 1538 / 2672}  & \cellcolor{ggreen!90}{13462 / 2697 / 2713}   & \cellcolor{ggreen!30}{9170 / 2883 / 2697}    & \cellcolor{ggray!90}{- / 267 / 2730}    \\
                    & \# Avg. Tokens & 492.8 / 47.3  & 521.3 / 55.4 & 500.6 / 71.8 & 516.8 / 49.4  & 524.2 / 48.4  & 458.3 / 54.3  \\ \hline
\multirow{2}{*}{Fr} & \# Samples   & \cellcolor{ggreen!90}{53232 / 5161 / 5258}   & \cellcolor{ggreen!90}{53232 / 5161 / 5258}   & \cellcolor{ggray!90}{- / 1449 / 2337}   & \cellcolor{ggreen!90}{10628 / 2605 / 2400}   & \cellcolor{ggreen!30}{7281 / 2750 / 2386}    & \cellcolor{ggray!90}{- / 232 / 2391}        \\
                    & \# Avg. Tokens & 659.4 / 45.3  & 659.3 / 55.5   & 617.3 / 73.1   & 649.0 / 48.5 & 673.4 / 47.3  & 589.9 / 54.4  \\ \hline
\multirow{2}{*}{Hi} & \# Samples   & \cellcolor{ggreen!30}{5707 / 1538 / 2672}    & \cellcolor{ggray!90}{- / 1449 / 2337}       & \cellcolor{ggreen!30}{5707 / 1538 / 2672}   & \cellcolor{ggray!90}{- / 1134 / 2000}       & \cellcolor{ggray!90}{- / 1266 / 2146}       & \cellcolor{ggray!90}{- / 180 / 2091}   \\
                    & \# Avg. Tokens & 682.1 / 46.2  & 668.3 / 58.2  & 684.3 / 72.3 & 637.9 / 50.5  & 626.1 / 48.7  & 627.4 / 53.0  \\ \hline
\multirow{2}{*}{Zh} & \# Samples   & \cellcolor{ggreen!90}{13462 / 2697 / 2713}   & \cellcolor{ggreen!90}{10628 / 2605 / 2400}   & \cellcolor{ggray!90}{- / 1134 / 2000}   & \cellcolor{ggreen!90}{13462 / 2697 / 2713}   & \cellcolor{ggray!90}{- / 2392 / 2218}       & \cellcolor{ggray!90}{- / 90 / 2147}       \\
                    & \# Avg. Tokens & 428.4 / 46.4 & 432.9 / 58.1 & 388.7 / 73.6 & 429.1 / 49.2 & 371.1 / 49.8 & 373.2 / 55.5  \\ \hline
\multirow{2}{*}{Th} & \# Samples   & \cellcolor{ggreen!30}{9170 / 2883 / 2697}    & \cellcolor{ggreen!30}{7281 / 2750 / 2386}    & \cellcolor{ggray!90}{- / 1266 / 2146}     & \cellcolor{ggray!90}{- / 2392 / 2218}       & \cellcolor{ggreen!30}{9170 / 2883 / 2697}    & \cellcolor{ggray!90}{- / 191 / 2172}     \\
                    & \# Avg. Tokens & 488.6 / 44.5  & 504.9 / 56.2 & 424.6 / 71.8  & 412.1 / 51.0 & 490.1 / 48.2  & 404.1 / 54.2 \\ \hline
\multirow{2}{*}{Tr} & \# Samples   & \cellcolor{ggray!90}{- / 267 / 2730}        & \cellcolor{ggray!90}{- / 232 / 2391}        & \cellcolor{ggray!90}{- / 180 / 2091}    & \cellcolor{ggray!90}{- / 90 / 2147}         & \cellcolor{ggray!90}{- / 191 / 2172}        & \cellcolor{ggray!90}{- / 267 / 2730}       \\
                    & \# Avg. Tokens & 465.1 / 47.5 & 472.4 / 60.0 & 468.1 / 72.8 & 456.9 / 52.7 & 449.1 / 49.8 & 465.1 / 54.3 \\ \toprule[1pt]
\end{tabular} 
}

\caption{Statistics of re-splitted WikiLingua. \textit{\# Samples} denotes the number of samples in training / validation / test set. \textit{\# Avg. Tokens} represents the average tokens in the documents and summaries, respectively. \colorbox{ggreen!90}{Green}, \colorbox{ggreen!30}{light green} and \colorbox{ggray!90}{gray} indicate the \colorbox{ggreen!90}{high-resource}, \colorbox{ggreen!30}{low-resource} and \colorbox{ggray!90}{zero-shot} directions, respectively.} 
\label{table:statistics}
\vspace{-0.3cm}
\end{table*}

\subsection{Task-Specific Pre-Training}
\label{subsec:task_spe_pretraining}
Task-specific pre-training aims to narrow the gap between the pre-training and fune-tuning stages. We directly adopt M2MS as its pre-training task. Grounding the truth that high-quality M2MS samples are difficult to collect, we construct the pseudo samples from multi-lingual unlabeled corpora.

In detail, for a source-language document $D=\{s^{src}_{i}\}^{|D|}_{i=1}$, where $s^{src}_{i}$ denotes the $i$-th sentence in $D$. Following previous monolingual pre-trained summarization methods~\cite{pmlr-v119-zhang20ae,Xiao2021PRIMERPM}, we calculate the importance of each sentence as $\mathcal{S}(s^{src}_{i})$ $=$ \textsc{Rouge-1}$(s^{src}_{i},D/s^{src}_{i})$, where $D/s^{src}_{i}$ indicates the rest of the document after $s^{src}_{i}$ is removed.
The sentences with high importance are selected as the gap sentences $S^{src}_{*}=\{s^{src}_{g_{i}}\}^{|S^{src}_{*}|}_{i=1}$ ($g_{i} \in \{1,2,...,|D|\}$), which are further translated to a different target language $S^{trg}_{*}=\{s^{trg}_{g_{i}}\}^{|S^{trg}_{*}|}_{i=1}$ via Google Translation\footnote{\url{https://cloud.google.com/translate}}. In this manner, the source-language document $D$ paired with source/target-language gap sentences $S^{src}_{*}$/$S^{trg}_{*}$ could constitute a pseudo pre-training sample.

\vspace{0.5ex}
\noindent \textbf{Quality Controlling.} Since machine translation results might contain flaws, we further employ \textit{round-trip translation} strategy as suggested by \citet{zhu-etal-2019-ncls} and \citet{feng-etal-2022-msamsum}. For each gap sentence $s^{src}_{g_{i}}$ in $D$, the translated counterpart $s^{trg}_{g_{i}}$ is translated back to the source language, which we denote as $s^{src'}_{g_{i}}$. If the \textsc{Rouge-1} score between $s^{src}_{g_{i}}$ and $s^{src'}_{g_{i}}$ is less than the pre-defined threshold $\lambda$, the corresponding pseudo sample will be discarded.

\vspace{0.5ex}
\noindent \textbf{Input Format.} To help the model trade off between (1) generating new sentences instead of translating part of input sentences, and (2) learning the translation pattern\footnote{In CLS, \citet{zhu-etal-2019-ncls} find some words in summaries are directly translated from the source words.}~\cite{zhu-etal-2020-attend}, half of source-language gap sentences in $D$ are randomly masked with a special token \texttt{<mask-sent>}.\footnote{We also attempt to mask all gap sentences or do not mask any gap sentences, the results underperform that of masking half of the gap sentences.}

\section{Experiments}

\subsection{Benchmark Datasets}
\label{subsec:datasets_and_metrics}
In order to evaluate M2MS models, two requirements should be met in datasets, \emph{i.e.}, (1) involving multiple languages and summarization directions, and (2) having abundant samples in each direction.
Thus, we choose WikiLingua~\cite{ladhak-etal-2020-wikilingua} and CrossSum~\cite{Hasan2021CrossSumBE}.

\begin{table*}[t]
\centering
\resizebox{0.87\textwidth}{!}
{
\begin{tabular}{r|cccccc||ccc}
\bottomrule[1pt]
\multicolumn{10}{c}{Non-Trivial Zero-Shot Directions}                                                                 \\ \hline
\multicolumn{1}{r|}{\multirow{2}{*}{Direction}} & \multicolumn{5}{c}{Tr$\Rightarrow$Others} & \multicolumn{1}{c||}{\cellcolor{ggray!90}{}} & \multicolumn{3}{c}{Any$\Rightarrow$Tr} \\
 & Tr$\Rightarrow$En       & Tr$\Rightarrow$Fr       & Tr$\Rightarrow$Hi       & Tr$\Rightarrow$Zh       & Tr$\Rightarrow$Th       & \multicolumn{1}{c||}{\cellcolor{ggray!90}{\multirow{-2}{*}{Avg.}}}  & En$\Rightarrow$Tr & Fr$\Rightarrow$Tr & Tr$\Rightarrow$Tr       \\ \hline
mT5 (580M)     & 9.5 / 61.6 & 10.0 / 63.8 & 7.8 / 59.1  & 12.6 / 59.6 & 14.0 / 59.6 & \cellcolor{ggray!90}{10.8 / 60.7} & 2.2 / 48.9 & 2.1 / 48.8 & 2.0 / 48.2 \\
mBART (610M)     & 10.6 / 62.1 & 10.8 / 64.4 & 9.9 / 60.1  & 13.5 / 60.0 & 15.2 / 59.9 & \cellcolor{ggray!90}{12.0 / 61.3} & 2.1 / 49.0 & 2.1 / 49.2 & 2.1 / 48.4  \\
\textsc{Pisces} (610M)    & 20.2 / 68.2 & 19.6 / 68.9 & 15.7 / 64.9 & 21.2 / 66.7 & 22.9 / 64.9 & \cellcolor{ggray!90}{19.9 / 66.7} & 3.1 / 53.8 & 2.8 / 53.4 & 3.7 / 52.9 \\ \toprule[1pt]
\end{tabular}
}

\resizebox{0.90\textwidth}{!}
{
\begin{tabular}{r|ccccccccc}
\bottomrule[1pt]
\multicolumn{10}{c}{Conventional Zero-Shot Directions}                                                                                                          \\ \hline
Direction & Fr$\Rightarrow$Hi       & Hi$\Rightarrow$Fr       & Hi$\Rightarrow$Zh       & Zh$\Rightarrow$Hi       & Hi$\Rightarrow$Th       & Th$\Rightarrow$Hi       & Zh$\Rightarrow$Th       & Th$\Rightarrow$Zh       & \cellcolor{ggray!90}{Avg.}  \\ \hline
mT5 (580M)     & 18.8 / 66.9 & 23.0 / 71.2 & 23.1 / 68.3 & 17.5 / 66.1 & 25.8 / 65.9 & 17.8 / 66.4 &  27.2 / 66.9  & 24.6 / 69.2 & \cellcolor{ggray!90}{22.2 / 67.6} \\
mBART (610M)    & 19.6 / 68.1 & 23.6 / 72.1 & 24.0 / 69.1 & 18.1 / 66.9 & 26.7 / 67.4 & 18.8 / 67.4 & 27.8/ 67.6  & 25.0 / 69.4 & \cellcolor{ggray!90}{23.0 / 68.5} \\
\textsc{Pisces} (610M)    & 21.4 / 69.1 & 26.1 / 72.9 & 26.1 / 70.4 & 20.3 / 68.5 & 29.1 / 68.5 & 21.4 / 69.0 & 29.9 / 68.9 & 27.0 / 71.0 & \cellcolor{ggray!90}{25.2 / 69.8} \\ \toprule[1pt]
\end{tabular}
}

\resizebox{0.90\textwidth}{!}
{
\begin{tabular}{r|ccccccccc}
\bottomrule[1pt]
\multicolumn{10}{c}{Low-Resource Directions}                                                                                                       \\ \hline
Direction & Hi$\Rightarrow$Hi       & Th$\Rightarrow$Th       & En$\Rightarrow$Hi       & Hi$\Rightarrow$En       & En$\Rightarrow$Th       & Th$\Rightarrow$En       & Fr$\Rightarrow$Th       & Th$\Rightarrow$Fr       & \cellcolor{ggray!90}{Avg.}        \\ \hline
mT5 (580M)     & 24.7 / 70.7 & 32.5 / 69.6 & 20.8 / 68.5 & 27.1 / 72.3 & 29.9 / 68.3 & 27.8 / 73.1 & 28.1 / 67.3 & 25.3 / 72.2 & \cellcolor{ggray!90}{27.0 / 70.2} \\
mBART (610M)     & 25.3 / 71.1 & 33.1 / 70.1 & 21.9 / 69.3 & 27.8 / 72.8 & 30.7 / 69.1 & 28.6 / 73.4 & 29.0 / 68.5 & 26.0 / 72.9 & \cellcolor{ggray!90}{27.8 / 70.9} \\
\textsc{Pisces} (610M)   & 26.5 / 71.8 & 34.2 / 70.7 & 23.7 / 70.3 & 29.5 / 73.6 & 31.9 / 70.1 & 30.0 / 74.0 & 30.0 / 69.2 & 27.4 / 73.8 & \cellcolor{ggray!90}{29.2 / 71.7} \\ \toprule[1pt]
\end{tabular}
}

\resizebox{1.00\textwidth}{!}
{
\begin{tabular}{r|cccccccccc}
\bottomrule[1pt]
\multicolumn{11}{c}{High-Resource Directions}                                                                                                                    \\ \hline
Direction & En$\Rightarrow$En       & Fr$\Rightarrow$Fr       & Zh$\Rightarrow$Zh       & En$\Rightarrow$Fr       & Fr$\Rightarrow$En       & En$\Rightarrow$Zh       & Zh$\Rightarrow$En       & Fr$\Rightarrow$Zh       & Zh$\Rightarrow$Fr       & \cellcolor{ggray!90}{Avg.}        \\ \hline
mT5 (580M)    & 30.9 / 74.0 & 29.8 / 74.3 & 31.0 / 72.5 & 26.8 / 73.3 & 27.5 / 72.7 & 27.5 / 70.9 & 28.0 / 73.0 & 26.4 / 70.3 & 25.5 / 72.5 & \cellcolor{ggray!90}{28.2 / 72.6} \\
mBART (610M)     & 31.7 / 74.6 & 29.7 / 74.7 & 31.5 / 72.7 & 27.8 / 73.9 & 28.3 / 73.4 & 28.2 / 71.5 & 28.8 / 73.4 & 27.1 / 70.9 & 26.5 / 73.3 & \cellcolor{ggray!90}{28.8 / 73.2} \\
\textsc{Pisces} (610M)    & 32.4 / 75.0 & 30.3 / 75.0 & 32.1 / 73.0 & 28.5 / 74.3 & 29.0 / 73.8 & 28.8 / 71.9 & 29.7 / 73.9 & 27.4 / 71.3 & 27.6 / 73.7 & \cellcolor{ggray!90}{29.5 / 73.5} \\ \toprule[1pt]
\end{tabular}

}

\caption{Experimental results on WikiLingua. \colorbox{ggray!90}{Avg.} indicates the average score for each cluster of directions. \textsc{Pisces} is significantly better than mBART with t-test p < 0.01 in all directions.} 
\label{table:main_results}
\vspace{-0.3cm}
\end{table*}

The original WikiLingua dataset, which involves 18 languages, is designed for CLS task. The 18 languages constitute 306 (18$\times$17) cross-lingual directions, each of which contains about 18k CLS samples in average. For each document, WikiLingua also contains its summary in the original language. Therefore, the dataset could be used to evaluate M2MS models. However, the original splitting is for CLS. Thus, we re-split WikiLingua with the special consideration for M2MS:
for each document in the test (or validation) set of one direction, the document and its parallel documents\footnote{For each document, WikiLingua usually contains its parallel documents in other languages.} are not allowed to appear in the training and validation (or test) sets of other directions.
This rule reduces the likelihood that learning shortcuts.
We also intentionally create several zero-shot directions.

We focus on six languages in this work: English (En), Chinese (Zh), French (Fr), Hindi (Hi), Turkish (Tr) and Thai (Th). After re-splitting, the statistics are shown in Table~\ref{table:statistics}. There are \textbf{9 high-resource directions} each of which contains more than 10k training samples. The other \textbf{8 directions} with less than 10k training samples are considered as \textbf{low-resource directions}. The remaining 19 zero-shot directions have no training sample.
According to \emph{whether both the source and target languages appear in the whole training set}, we further divide them into \textbf{11 non-trivial and 8 conventional zero-shot directions}.
Note that Tr never appears in the training set of any direction, thus, in other words, the non-trivial zero-shot directions involve Tr while the conventional counterparts do not.
We call Tr an \emph{unseen language}.
Though there is no training data in a conventional zero-shot direction, both its source and target languages might have training data with a pivot language, making it less challenging than the non-trivial ones.
Taking the conventional zero-shot direction Hi$\Rightarrow$Zh as an example, the training data in Hi$\Rightarrow$En and En$\Rightarrow$Zh could bridge the gap between Hi and Zh.
For statistics of the CrossSum dataset used in our experiments, please refer to Appendix~\ref{appendix:crosssum1}.

\subsection{Experimental Setup}

\noindent \textbf{Baselines.} We use mBART-50~\cite{Tang2020MultilingualTW} and mT5~\cite{xue-etal-2021-mt5} as baselines, which have achieved state-of-the-art performances on many CLS/MLS datasets~\cite{perez-beltrachini-lapata-2021-models,Hasan2021CrossSumBE,feng-etal-2022-msamsum}.

\vspace{0.5ex}
\noindent \textbf{Metrics.} We adopt \textsc{Rouge-1/2/l}~\cite{Lin2004ROUGEAP} and \textsc{BertScore}~\cite{Zhang2020BERTScoreET} in our experiments. The \textsc{Rouge} scores measure the lexical overlap between the generated summaries and corresponding references, while the \textsc{BertScore} measures the semantic similarity. These metrics are calculated by \textit{multi-lingual rouge}\footnote{\url{https://github.com/csebuetnlp/xl-sum/tree/master/multilingual_rouge_scoring}} and \textit{bert-score}\footnote{\url{https://github.com/Tiiiger/bert_score}} toolkits, respectively. The \textsc{BertScore} is based on \textit{bert-base-multilingual-cased} model.
The statistical significance test~\cite{koehn-2004-statistical} is also employed for a fair comparison.

\vspace{0.5ex}
\noindent \textbf{Implementation Details.} The implementation details of the pre-training objectives, pre-training corpora and fine-tuning hyper-parameters are given in Appendix~\ref{appendix:implementation}.

\subsection{Quantitative Results}
\label{subsec:results}
Table~\ref{table:main_results} shows the results on WikiLingua in terms of average \textsc{Rouge} score (\textsc{Rs}) and \textsc{BertScore} (\textsc{Bs}). 
Full results on \textsc{Rouge-1/2/l} are given in Appendix~\ref{appendix:rouge}.
The experimental results on CrossSum also verify the superiority of \textsc{Pisces}, which are provided in Appendix~\ref{appendix:crosssum2}.

\vspace{0.5ex}
\noindent \textbf{\textsc{Pisces} vs. Baselines.} Our \textsc{Pisces} outperforms mBART-50 and mT5 in all directions, indicating its superiority. Specifically, \textsc{Pisces} achieves an average increase of 7.9 \textsc{Rs} and 5.4 \textsc{Bs} over mBART-50 in non-trivial zero-shot directions when the target language is not Tr.
Compared with mBART-50, the average improvement in conventional zero-shot directions is 2.2 \textsc{Rs} / 1.3 \textsc{Bs}, while the counterpart in low-resource directions is 1.4 \textsc{Rs} / 0.8 \textsc{Bs}. As for high-resource directions, \textsc{Pisces} outperforms mBART-50 by 0.7 \textsc{Rs} and 0.3 \textsc{Bs} in average. It is not difficult to find that the fewer resources in a direction, the greater the improvement brought by our \textsc{Pisces}. This finding also indicates the potentiality of our model when faced with the real-world scenario, since there are thousands of languages in the world and most directions are low-resource or zero-shot. Through the cross-lingual and task-specific pre-training stages, \textsc{Pisces} facilitates the transfer of task knowledge from high-resource directions to the low-resource and zero-shot ones.

\vspace{0.5ex}
\noindent \textbf{Non-Trivial Zero-Shot Direction.} As shown in Table~\ref{table:main_results}, we divide the non-trivial zero-shot directions into two categories (\emph{i.e.}, Tr$\Rightarrow$Others and Any$\Rightarrow$Tr) according to whether Tr is the target language.
We discover that the results in Any$\Rightarrow$Tr directions\footnote{Results on Hi/Zh/Th$\Rightarrow$Tr are given in Appendix~\ref{appendix:rouge}} are significantly worse than the Tr$\Rightarrow$Others counterparts.
This finding suggests that generating summaries in \emph{unseen languages} is more difficult than understanding documents in \emph{unseen languages}.
This is because the encoder could partly understand the \emph{unseen languages} through the shared vocabulary and the similar syntax constituent with other languages.
But for the decoder, we only change its language tag to expect it can generate summaries in \emph{unseen languages}.
This requires the decoder to \emph{simultaneously} (1) capture the relationships between the unseen language tag and the unseen language tokens and (2) summarize documents. However, the pre-trained model only meets the requirement (1) in the pre-training stage\footnote{Though \textsc{Pisces} has been pre-trained with pseudo M2MS samples, there is still a large gap between the pseudo samples and downstream samples, \emph{e.g.}, text style and domain.}, while requirement (2) in the fine-tuning stage, making it hard to simultaneously meet both requirements, and consequently, cannot generate summaries in unseen languages.
We reserve this challenge for future work.

\begin{table}[t]
\centering
\resizebox{0.48\textwidth}{!}
{
\begin{tabular}{lcccc}
\bottomrule[1pt]
             & Fr$\Rightarrow$Hi       & Hi$\Rightarrow$Fr       & Hi$\Rightarrow$Zh       & Zh$\Rightarrow$Hi       \\ \toprule[1pt]
\textsc{Pisces}       & \textbf{21.4} / \textbf{69.1} & \textbf{26.1} / \textbf{72.9} & \textbf{26.1} / \textbf{70.4} & \textbf{20.3} / \textbf{68.5} \\
\quad w/o TS & 20.7 / 68.6 & 25.2 / 72.8 & 25.1 / 69.9 & 19.5 / 67.9 \\
\quad w/o CL & 20.6 / 68.8 & 25.2 / \textbf{72.9} & 25.3 / 70.0 & 19.5 / 67.8 \\
\quad w/o TS \& CL   & 19.6 / 68.1 & 23.6 / 72.1 & 24.0 / 69.1 & 18.1 / 66.9 \\ \bottomrule[1pt]
\end{tabular}
}

\caption{Results of ablation studies.} 
\label{table:ablations}
\vspace{-0.3cm}
\end{table}

\begin{table}[t]
\centering
\resizebox{0.48\textwidth}{!}
{
\begin{tabular}{l|ccccccccc}
\toprule[1pt]
\multicolumn{1}{c|}{\multirow{3}{*}{Model}} & \multicolumn{9}{c}{WikiLingua}                                                                                \\ \cline{2-10} 
                       & \multicolumn{3}{c|}{En$\Rightarrow$Zh}              & \multicolumn{3}{c|}{Zh$\Rightarrow$En}              & \multicolumn{3}{c}{En$\Rightarrow$En} \\
                       & IF   & CC   & \multicolumn{1}{c|}{GM}   & IF   & CC   & \multicolumn{1}{c|}{GM}   & IF      & CC     & GM     \\ \midrule[1pt]
mT5                    & 2.93 & 3.12 & \multicolumn{1}{c|}{3.06} & 2.98 & 3.29 & \multicolumn{1}{c|}{3.03} & 3.16    & 3.84   & 3.52   \\
mBART               & 3.09 & 3.38 & \multicolumn{1}{c|}{3.14} & 3.15 & 3.53 & \multicolumn{1}{c|}{3.26} & 3.27    & 3.96   & 3.71   \\
\textsc{Pisces}                 & \textbf{3.17} & \textbf{3.56} & \multicolumn{1}{c|}{\textbf{3.41}} & \textbf{3.24} & \textbf{3.76} & \multicolumn{1}{c|}{\textbf{3.54}} & \textbf{3.52}    & \textbf{4.28}   & \textbf{4.16}   \\ \bottomrule[1pt]
\end{tabular}
}

\caption{Human evaluation results. ``IF'', ``CC'' and ``GM'' denote informativeness, conciseness and grammaticality, respectively.}
\label{table:human_evaluation}
\vspace{-0.3cm}
\end{table}

\vspace{0.5ex}
\noindent \textbf{Ablations.} We conduct ablation studies to investigate the effect of the cross-lingual and task-specific pre-training stages.
We run the following ablations:
\begin{itemize}[leftmargin=*,topsep=0pt]
\setlength{\itemsep}{0pt}
\setlength{\parsep}{0pt}
\setlength{\parskip}{0pt}
\item \textbf{\textsc{Pisces} w/o TS}. To demonstrate the effectiveness of the task-specific pre-training, we also pre-train a variant \textsc{Pisces} model which does not include the task-specific pre-training stage.
\item \textbf{\textsc{Pisces} w/o CL}. To measure the effectiveness of the cross-lingual pre-training, we remove this stage in the whole pre-training process, resulting in another variant \textsc{Pisces}.
\item \textbf{\textsc{Pisces} w/o TS \& CL} removes both the cross-lingual and task-specific pre-training stages, which is the same as mBART-50.
\end{itemize}

As shown in Table~\ref{table:ablations}, we conduct ablation studies in several conventional zero-shot directions (results in more directions are provided in Appendix~\ref{appendix:ablations}). In each case, the \textsc{Rs} and \textsc{Bs} are lower than vanilla \textsc{Pisces}. In addition, both \textsc{Pisces} w/o TS and \textsc{Pisces} w/o CL outperform \textsc{Pisces} w/o TS \& CL. Thus, the effectiveness of both stages is proved.

\begin{table}[t]
\centering
\resizebox{0.50\textwidth}{!}
{
\begin{tabular}{cl}
\toprule[1pt]
\multicolumn{2}{c}{\textbf{How to Download Photos from Your iPhone to a Computer}}  \\ \bottomrule[1pt]
\tabincell{c}{Turkish\\Document}          & \tabincell{l}{iPhone'un şarj kablosunun bir ucunu iPhone'un şarj girişine tak, ardından \\USB ucunu bilgisayarının USB girişlerinden birine tak. Kilidini açmak i-\\çin parolanı (veya TouchID'ni ya da FaceID'ni) gir ve iPhone'undaki Ho-\\me düğmesine bas. Devam etmeden önce, istenirse "Bu bilgisayara güve-\\nilsin mi?" kısmında Güven seçeneğine dokun. Mac'in Dock'unda çok \\renkli bir çarkıfeleğe benzeyen Fotoğraflar uygulaması simgesine tıkla.  \\Fotoğraflar uygulaması iPhone'unu bağladığında otomatik olarak açılabilir. \\iPhone'un simgesi, uygulamanın penceresinin sol üst köşesinde görünmeli-\\dir. Fotoğrafların alınıp içeri aktarılacağı yer olarak pencerenin sol tarafında \\iPhone'unun adına tıkla. Bunu penceredeki resimlere tıklayarak yap. Bilgi-\\sayarında olmayan tüm fotoğrafları içeri aktarmak istiyorsan bu adımı atla. \\Bu, pencerenin sağ üst köşesindedir. Seçtiğin fotoğraf sayısı bu butonda \\görünecektir (örneğin, 5 Seçileni İçeri Aktar). iPhone'undaki Mac bilgi-\\sayarında olmayan tüm fotoğrafları aktarmak istiyorsan Tüm Yeni Ögeleri \\İçeri Aktar seçeneğine tıkla. Bu, pencerenin sol tarafındadır. Az önce akt-\\ardığın fotoğraflar bu sayfada listelenir.}                             \\ \hline
mBART & \tabincell{l}{Examine the iphone's keyboard. Click the "screen" button to view the photos. \\Click the "screen" button to view the list of available photos.} \\ \hline
\textsc{Pisces}                    & \tabincell{l}{\textbf{Connect your iphone to} computer. \textbf{Unlock your iphone.} Click \textbf{the "photos"} \\\textbf{app}. \textbf{Select the photos you} wish \textbf{to download.} Click the "choose photos" opt-\\ion. \textbf{Select the photos you} wish \textbf{to download.} Click the "download" button.}  \\ \hline
\tabincell{c}{Ground\\Truth}              & \tabincell{l}{\textbf{Connect your iphone to} your mac. \textbf{Unlock your iphone.} Open \textbf{the photos} \\ \textbf{app.} Select your iphone. \textbf{Select the photos you}'d like \textbf{to download.} Click \\import selected. Click imports.}   \\ \bottomrule[1pt]
\end{tabular}
}

\caption{An example of Tr$\Rightarrow$En summarization.} 
\label{table:full_case_study}
\vspace{-0.3cm}
\end{table}

\subsection{Qualitative Results}

\noindent \textbf{Human Evaluation.} Following~\citet{zhu-etal-2020-attend,Liang2022AVH}, we conduct the human evaluation on 50 random samples extracted from WikiLingua (En$\Rightarrow$Zh, Zh$\Rightarrow$En and En$\Rightarrow$En, respectively). Three graduate students are invited to assess the generated summaries from three aspects: informativeness (IF), conciseness (CC) and grammaticality (GM). The scoring adopts a 5-point scale from 1 (worst) to 5 (best). Table~\ref{table:human_evaluation} shows the average results. The IF, CC and GM scores of \textsc{Pisces} are significantly better than those of mT5 or mBART-50, demonstrating the effectiveness of our model.

\vspace{0.5ex}
\noindent \textbf{Case Study.} Table~\ref{table:full_case_study} shows an example Turkish document, the generated summary and the ground truth summary.
Though the summary generated by \textsc{Pisces} contains a repeated sentence, it has good overlaps with the ground truth. But for mBART-50, the generated summary is not relevant to the core idea of the document.
This observation indicates that, through the cross-lingual and task-specific pre-training, \textsc{Pisces} could better transfer the task knowledge from high-resource directions to zero-shot ones, and even has the ability to generate summaries for the documents whose language does not occur in the fine-tuning stage.

\vspace{0.5ex}
\noindent \textbf{Error Analysis.} To further study how future research could advance M2MS, we take a closer look at the generation errors of \textsc{Pisces} and analyze them in Appendix~\ref{appendix:errors}.

\section{Conclusion}
In this paper, we unify MLS and CLS to M2MS.
Through carefully-designed preliminary studies, we discuss that unifying MLS and CLS to M2MS is valuable.
In addition, we propose \textsc{Pisces}, the first pre-trained M2MS model, which contains three pre-training stages to enable the model learn the multi-lingual language modeling, cross-lingual ability and summarization ability.
Extensive experiments show its superiority compared with the state-of-the-art baselines (mBART-50 and mT5). The case study further demonstrates that our model could even generate summaries for the documents whose language does not occur in the fine-tuning stage.

\section*{Ethical Considerations}
In this section, we consider potential ethical issues of our model.
In this paper, we propose \textsc{Pisces} which utilizes mBART-50~\cite{Tang2020MultilingualTW} as the meta pre-trained model and further suffers from the cross-lingual pre-training and task-specific pre-training stages.
The pre-training samples are constructed from OPUS~\cite{tiedemann-thottingal-2020-opus} and mC4~\cite{xue-etal-2021-mt5} corpora.
To construct the pseudo M2MS samples in the task-specific pre-training stage, Google Translation is also adopted to translate gap sentences.
Therefore, \textsc{Pisces} might involve the same biases and toxic behaviors exhibited by language models, pre-training corpora and Google Translation.

\section*{Limitations}
While we show that \textsc{Pisces} outperforms mBART-50 on WikiLingua~\cite{ladhak-etal-2020-wikilingua}, there are some limitations worth considering in future work:
(1) \textsc{Pisces} still struggles to generate summaries in unseen languages (Section~\ref{subsec:results});
(2) In this work, we focus on six languages in total, and future work could extend our method to more languages.

\section*{Acknowledgements}
This work is supported by the National Natural Science Foundation of China (No.62072323, 62102276), Shanghai Science and Technology Innovation Action Plan (No. 22511104700), the Natural Science Foundation of Jiangsu Province (Grant No. BK20210705), the Natural Science Foundation of Educational Commission of Jiangsu Province, China (Grant No. 21KJD520005) and the Priority Academic Program Development of Jiangsu Higher Education Institutions.

\bibliography{anthology}
\bibliographystyle{acl_natbib}

\appendix

\section{Word Embeddings of the Unseen Language and Other Languages}
\label{appendix:token_representation}

\begin{figure}[h]
\centering
\subfigure[mBART (\texttt{M2MS})]{
  \includegraphics[width=0.45\linewidth]{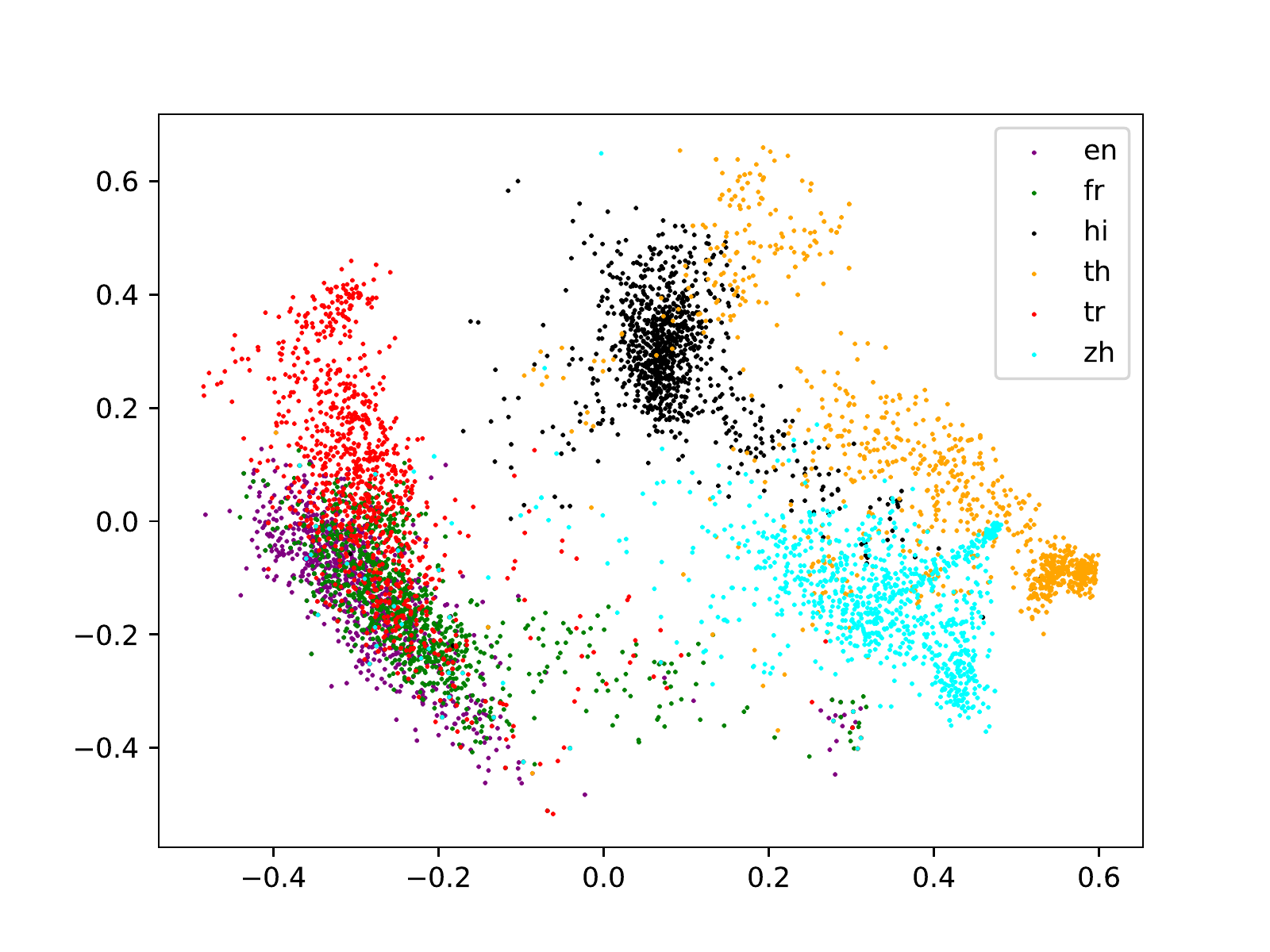}
}
\subfigure[mBART (\texttt{U-CLS})]{
  \includegraphics[width=0.45\linewidth]{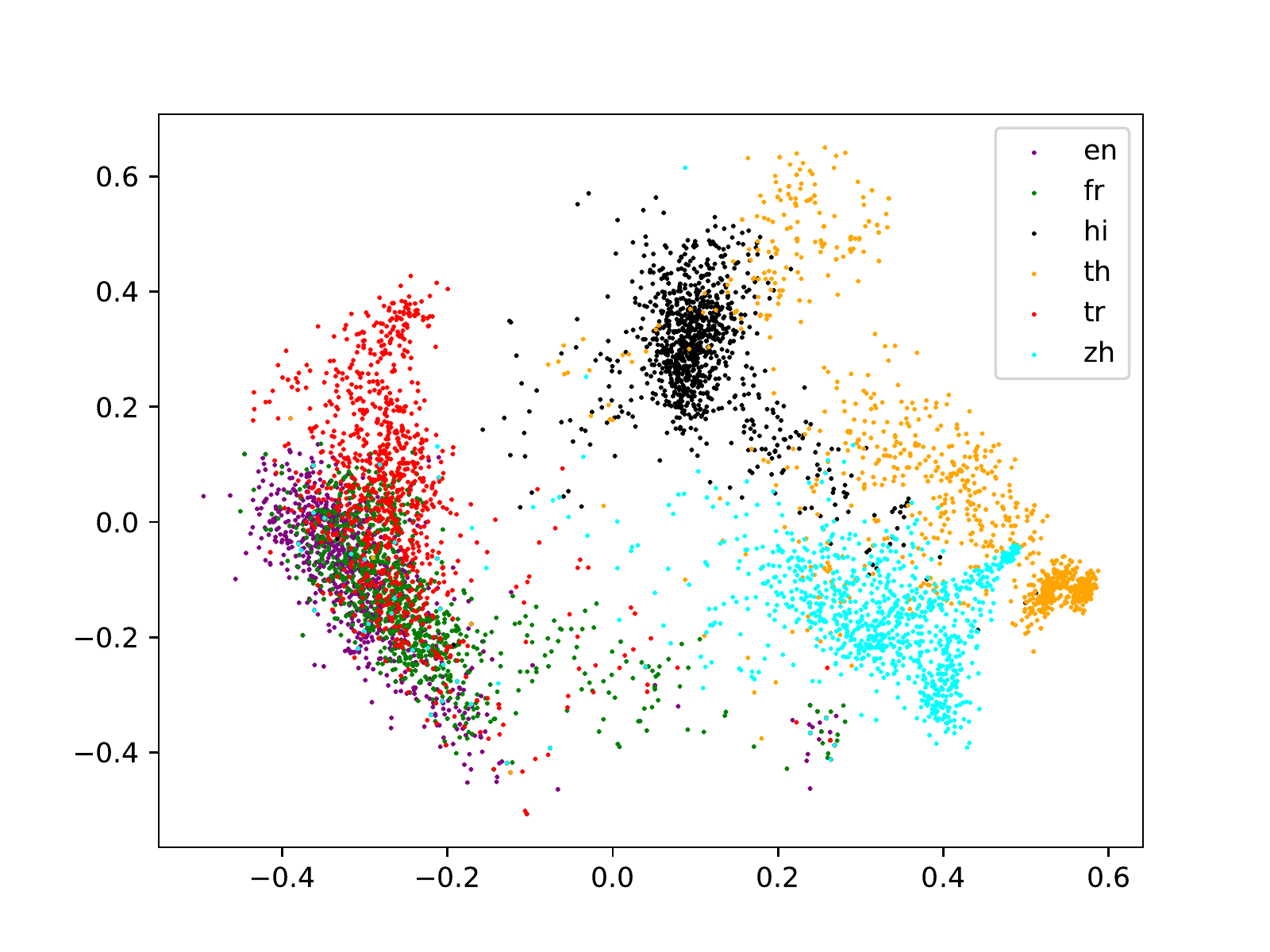}
}
\caption{Visualization of word embeddings from mBART (\texttt{M2MS}) and mBART (\texttt{U-CLS}). \textcolor{red}{Tr} is the unseen language.}
\label{fig:pca_visual_embedding}
\end{figure}

\begin{table*}[t]
\centering
\resizebox{0.90\textwidth}{!}
{
\begin{tabular}{c|rrrrrrrrr|r}
\toprule[1pt]
Direction & MultiUN & CCMatrix & CCAligned & MultiCCAligned & XLEnt   & Europarl & QED    & TED    & WMT   & \textbf{Sum}      \\ \bottomrule[1pt]
En$\Leftrightarrow$Fr     & -       & -        & -         & -              & -       & 349291   & 152623 & 77188  & 4648  & 583750   \\
En$\Leftrightarrow$Hi     & -       & 2959722  & -         & -              & 405366  & -        & 1211   & 9039   & 568   & 3375906  \\
En$\Leftrightarrow$Th     & -       & -        & 1947729   & -              & 246976  & -        & 52140  & 30765  & -     & 2277610  \\
En$\Leftrightarrow$Tr     & -       & -        & 2496997   & -              & 761750  & -        & 94212  & 72674  & 3819  & 3429452  \\
En$\Leftrightarrow$Zh     & -       & -        & -         & -              & 1258289 & -        & -      & 3158   & 3658  & 1265105  \\
Fr$\Leftrightarrow$Hi     & -       & -        & -         & 619040         & 97082   & -        & 660    & 8816   & -     & 725598   \\
Fr$\Leftrightarrow$Th     & -       & -        & -         & 737469         & 67292   & -        & 34418  & 30024  & -     & 869203   \\
Fr$\Leftrightarrow$Tr     & -       & -        & -         & 1321431        & 183282  & -        & 61412  & 69931  & -     & 1636056  \\
Fr$\Leftrightarrow$Zh     & 1494829 & -        & -         & -              & 211039  & -        & 2041   & 3088   & -     & 1710997  \\
Hi$\Leftrightarrow$Th     & -       & -        & -         & 436284         & 65870   & -        & 484    & 4526   & -     & 507164   \\
Hi$\Leftrightarrow$Tr     & -       & 1099853  & -         & -              & 111573  & -        & 544    & 8384   & -     & 1220354  \\
Hi$\Leftrightarrow$Zh     & -       & 445148   & -         & -              & 97732   & -        & 15     & 650    & -     & 543545   \\
Th$\Leftrightarrow$Tr     & -       & -        & -         & 617566         & 86156   & -        & 40026  & 29602  & -     & 773350   \\
Th$\Leftrightarrow$Zh     & -       & -        & -         & -              & 54637   & -        & 2390   & 2169   & -     & 59196    \\
Tr$\Leftrightarrow$Zh     & -       & 1435286  & -         & -              & 169774  & -        & 1885   & 3125   & -     & 1610070  \\ \toprule[1pt]
\textbf{Total}     & 1494829 & 5940009  & 4444726   & 3731790        & 3816818 & 349291   & 444061 & 353139 & 12693 & 20587356 \\ \bottomrule[1pt]
\end{tabular}
}
\caption{Statistics of the constructed cross-lingual pre-training samples. Each entry shows the number of samples for each language pair in the corresponding corpus.} 
\label{table:samples_cross_lingual_pre_training}
\end{table*}

\begin{table*}[t]
\centering
\resizebox{0.80\textwidth}{!}
{
\begin{tabular}{ccccccccccc}
\toprule[1pt]
En$\Leftrightarrow$Fr  & En$\Leftrightarrow$Hi  & En$\Leftrightarrow$Th  & En$\Leftrightarrow$Tr  & En$\Leftrightarrow$Zh & Fr$\Leftrightarrow$Hi  & Fr$\Leftrightarrow$Th  & Fr$\Leftrightarrow$Tr  & Fr$\Leftrightarrow$Zh  & Hi$\Leftrightarrow$Th  & Hi$\Leftrightarrow$Tr   \\
190916 & 190916 & 190916 & 190916 & 88636 & 188351 & 190916 & 190916 & 190916 & 158518 & 190578  \\ \midrule[1pt]
Hi$\Leftrightarrow$Zh  & Th$\Leftrightarrow$Tr  & Th$\Leftrightarrow$Zh  & Tr$\Leftrightarrow$Zh  & En$\Rightarrow$En & Fr$\Rightarrow$Fr  & Hi$\Rightarrow$Hi  & Th$\Rightarrow$Th  & Tr$\Rightarrow$Tr  & Zh$\Rightarrow$Zh  & \textbf{Total}     \\
172039 & 190916 & 24160  & 190916 & 95458 & 95458  & 95458  & 95458  & 95458  & 95458  & 3113274 \\ \bottomrule[1pt]
\end{tabular}
}
\caption{Statistics of the constructed task-specific pre-training samples.} 
\label{table:samples_task_specific_pre_training}
\vspace{-0.3cm}
\end{table*}

To verify the word embeddings of the unseen language drift away from those of other languages after adding the monolingual training data, based on MUSE dictionary, we choose top frequent 1000 English words and the words with the same meaning in other five languages (\emph{i.e.}, Fr, Hi, Zh, Th and Tr).
Then, we calculate the embeddings of these words based on mBART (\texttt{M2MS}) and mBART (\texttt{U-CLS}), respectively.
For the word that consists of multiple tokens, the word embedding is the average of embeddings of those tokens.
As shown in Figure~\ref{fig:pca_visual_embedding}, we utilize Principal Component Analysis (PCA) to visualize the word embeddings from mBART (\texttt{M2MS}) and mBART (\texttt{U-CLS}).
In the PCA space, we further calculate the central point of each language by averaging the word embeddings in the language.
Then, we find the average distance between the central point of Tr and other languages is 0.426 / 0.407 for mBART (\texttt{M2MS}) / mBART (\texttt{U-CLS}).
This distance in vanilla mBART-50~\cite{Tang2020MultilingualTW} is 0.398.
Therefore, the monolingual training data used in mBART (\texttt{M2MS}) makes the word embeddings of the unseen language drift away from those of other languages.

\section{Implementation Details}
\label{appendix:implementation}

\noindent \textbf{Pre-Training Details.} We use mBART-50~\cite{Tang2020MultilingualTW} as the meta pre-trained model, and futher pre-train it via cross-lingual and task-specific pre-training stages.
The implementation of mBART-50 is based on the Transformers~\cite{wolf-etal-2020-transformers} library with default settings (12 encoder layers, 12 decoder layers and 1024 hidden states).
In cross-lingual pre-training, we dynamically mask 0-15\% tokens in the source-language sentences, and construct 20.6M samples from OPUS parallel corpora~\cite{tiedemann-thottingal-2020-opus}. In task-specific pre-training, we construct 3.1M training samples from mC4 corpus~\cite{xue-etal-2021-mt5}.
We set the total length of gap sentences to $k$\% of the document length, and $k$ is dynamically selected from $[5,10,15]$. The pre-defined $\lambda$ in the round-trip translation is 0.7. All experimental results listed in this paper are the average of 3 runs.

Table~\ref{table:samples_cross_lingual_pre_training} and Table~\ref{table:samples_task_specific_pre_training} show the statistics of the constructed samples in the cross-lingual pre-training and task-specific pre-training stages, respectively.
The cross-lingual pre-training and task-specific pre-training stages are conducted on 8 NVIDIA Tesla V100 GPUs with 32GB memory.
In the cross-lingual pre-training stage, we pre-train the model for 150K steps, with early stopping, 32 batch size, 3e-5 learning rate following~\citet{Xiao2021PRIMERPM} and 10K warmup steps.
In the task-specific pre-training stage, we pre-train the model for 100K steps, with early stopping, 4 batch size, 3e-5 learning rate and 10K warmup steps.

\begin{table*}[t]
\centering
\resizebox{0.90\textwidth}{!}
{
\begin{tabular}{clcccccc}
\midrule[1pt]
\diagbox[dir=NW]{Src}{Trg}                    &                & En                    & Fr                    & Hi        &  Zh & Th & Tr            \\ \bottomrule[1pt]
\multirow{2}{*}{En} & \# Samples              & 8000 / 1000 / 1000 & 1513 / 188 / 188   & 3784 / 463 / 481   & 3981 / 497 / 497   & 816 / 102 / 102  & 4542 / 568 / 566   \\
                    & \# Avg. Tokens          & 638.7 / 30.6       & 1013.0 / 43.2      & 899.9 / 41.0       & 914.6 / 35.5       & 1058.3 / 51.3    & 880.8 / 37.6       \\ \hline
\multirow{2}{*}{Fr} & \# Samples              & 1513 / 188 / 188   & 8000 / 1000 / 1000 & \cellcolor{ggray!90}{- / 308 / 308}      & \cellcolor{ggray!90}{- / 174 / 174}      & \cellcolor{ggray!90}{- / 92 / 93}      & \cellcolor{ggray!90}{- / 414 / 415}      \\
                    & \# Avg. Tokens          & 1124.3 / 33.7      & 710.9 / 40.8       & 1048.5 / 40.7      & 1358.3 / 37.6      & 1501.7 / 47.9    & 1058.3 / 38.9      \\ \hline
\multirow{2}{*}{Hi} & \# Samples              & 3784 / 463 / 481   & \cellcolor{ggray!90}{- / 308 / 308}      & 8000 / 1000 / 1000 & 1107 / 135 / 137   & \cellcolor{ggray!90}{- / 189 / 189}    & 2956 / 369 / 369   \\
                    & \# Avg. Tokens          & 862.0 / 31.6       & 1106.5 / 39.1      & 775.4 / 40.2       & 804.3 / 33.6       & 1186.3 / 49.4    & 712.8 / 34.3       \\ \hline
\multirow{2}{*}{Zh} & \# Samples              & 3981 / 497 / 497   & \cellcolor{ggray!90}{- / 174 / 174}      & 1107 / 135 / 137   & 8000 / 1000 / 1000 & \cellcolor{ggray!90}{- / 134 / 135}    & 1209 / 151 / 151   \\
                    & \# Avg. Tokens          & 725.0 / 32.7       & 1082.0 / 41.9      & 690.7 / 41.9       & 768.0 / 40.4       & 1059.7 / 52.7    & 642.4 / 36.6       \\ \hline
\multirow{2}{*}{Th} & \# Samples              & 816 / 102 / 102    & \cellcolor{ggray!90}{- / 92 / 93}        & \cellcolor{ggray!90}{- / 189 / 189}      & \cellcolor{ggray!90}{- / 134 / 135}      & 6616 / 826 / 826 & \cellcolor{ggray!90}{- / 238 / 239}      \\
                    & \# Avg. Tokens          & 957.1 / 34.5       & 1095.2 / 40.6      & 985.3 / 42.0       & 1036.3 / 38.7      & 1055.5 / 62.1    & 912.2 / 39.7       \\ \hline
\multirow{2}{*}{Tr} & \# Samples              & 4542 / 568 / 566   & \cellcolor{ggray!90}{- / 414 / 415}      & 2956 / 369 / 369   & 1209 / 151 / 151   & \cellcolor{ggray!90}{- / 238 / 239}    & 8000 / 1000 / 1000 \\
                    & \# Avg. Tokens          & 619.4 / 31.8       & 775.2 / 41.2       & 579.3 / 39.0       & 591.5 / 34.4       & 762.7 / 53.2     & 704.9 / 40.2  \\ \toprule[1pt]
\end{tabular} 
}

\caption{Statistics of CrossSum used in our experiments. \textit{\# Samples} denotes the number of samples in training / validation / test set. \textit{\# Avg. Tokens} represents the average tokens in the documents and summaries, respectively. \colorbox{ggray!90}{gray} indicates the \colorbox{ggray!90}{zero-shot} directions.}
\label{table:crosssum_statistics}
\vspace{-0.3cm}
\end{table*}

\begin{table*}[t]
\centering
\resizebox{0.98\textwidth}{!}
{
\begin{tabular}{c|rcccccc|c}
\toprule[1pt]
\multicolumn{1}{l|}{\diagbox[dir=NW]{Src}{Trg}} & \multicolumn{1}{c}{Model}  & En                        & Fr                        & Hi                        & Zh                        & Th                        & Tr                        & \multicolumn{1}{c}{Avg.} \\ \bottomrule[1pt]
\multirow{3}{*}{En}   & mT5 (580M)                & 30.1 / 8.3 / 22.3 / 66.3  & 30.7 / 10.4 / 22.8 / 66.2 & 30.2 / 8.9 / 24.8 / 67.4  & 26.1 / 6.6 / 22.6 / 65.8  & 27.4 / 8.6 / 21.8 / 60.1  & 25.8 / 9.9 / 22.5 / 66.7  & 28.4 / 8.8 / 22.8 / 65.4  \\
                      & mBART (610M)              & 31.2 / 8.7 / 22.8 / 66.9  & 32.8 / 12.4 / 24.5 / 66.9 & 32.6 / 9.5 / 25.5 / 68.3  & 29.6 / 8.2 / 24.3 / 67.0  & 30.8 / 10.4 / 23.4 / 62.9 & 26.3 / 10.2 / 22.8 / 67.2 & 30.6 / 9.9 / 23.9 / 66.5  \\
                      & Pisces (610M)             & 32.0 / 9.1 / 23.7 / 67.4  & 33.6 / 13.4 / 25.6 / 67.6 & 33.4 / 10.5 / 26.4 / 68.9 & 30.5 / 8.6 / 24.9 / 67.5  & 31.3 / 11.7 / 24.2 / 64.1 & 27.1 / 10.9 / 23.5 / 67.5 & 31.3 / 10.7 / 24.7 / 67.2 \\ \hline
\multirow{3}{*}{Fr}   & mT5 (580M)                & 30.7 / 10.1 / 23.4 / 66.6 & 30.8 / 11.9 / 23.5 / 66.2 & \cellcolor{ggray!90}{33.0 / 12.1 / 27.3 / 68.1} & \cellcolor{ggray!90}{39.3 / 21.5 / 33.1 / 69.9} & \cellcolor{ggray!90}{35.6 / 15.9 / 29.2 / 64.6} & \cellcolor{ggray!90}{25.3 / 10.7 / 22.3 / 65.8} & 32.5 / 13.7 / 26.5 / 66.9 \\
                      & mBART (610M)              & 31.9 / 10.3 / 23.8 / 67.4 & 32.0 / 12.9 / 24.3 / 66.6 & \cellcolor{ggray!90}{36.0 / 16.6 / 30.2 / 69.8} & \cellcolor{ggray!90}{41.3 / 23.4 / 36.7 / 70.9} & \cellcolor{ggray!90}{37.1 / 17.4 / 30.8 / 65.3} & \cellcolor{ggray!90}{29.5 / 14.1 / 26.1 / 68.2} & 34.6 / 15.8 / 28.7 / 68.0 \\
                      & Pisces (610M)             & 33.5 / 11.7 / 25.8 / 68.2 & 32.7 / 13.4 / 25.0 / 67.1 & \cellcolor{ggray!90}{39.7 / 19.5 / 33.9 / 71.5} & \cellcolor{ggray!90}{43.8 / 25.7 / 38.7 / 73.0} & \cellcolor{ggray!90}{42.8 / 25.3 / 35.7 / 69.2} & \cellcolor{ggray!90}{33.9 / 18.9 / 30.4 / 70.1} & 37.7 / 19.1 / 31.6 / 69.9 \\ \hline
\multirow{3}{*}{Hi}   & mT5 (580M)                & 29.7 / 9.3 / 23.2 / 67.2  & \cellcolor{ggray!90}{29.6 / 10.8 / 23.3 / 66.1} & 32.0 / 11.3 / 25.7 / 67.4 & 28.6 / 8.3 / 24.0 / 66.3  & \cellcolor{ggray!90}{29.8 / 11.0 / 23.8 / 62.6} & 22.0 / 7.4 / 19.8 / 65.5  & 28.6 / 9.7 / 23.3 / 65.9  \\
                      & mBART (610M)              & 31.5 / 9.9 / 24.0 / 67.7  & \cellcolor{ggray!90}{32.5 / 13.3 / 25.5 / 67.4} & 32.9 / 11.8 / 26.0 / 67.7 & 29.4 / 8.9 / 24.6 / 66.9  & \cellcolor{ggray!90}{33.7 / 15.1 / 27.6 / 65.0} & 22.6 / 7.8 / 19.7 / 65.7  & 30.4 / 11.1 / 24.6 / 66.7 \\
                      & Pisces (610M)             & 31.8 / 9.9 / 24.1 / 68.0  & \cellcolor{ggray!90}{35.3 / 15.9 / 28.0 / 68.7} & 33.8 / 12.5 / 26.8 / 68.3 & 32.5 / 10.8 / 27.3 / 68.9 & \cellcolor{ggray!90}{38.3 / 19.0 / 31.3 / 67.0} & 23.9 / 8.6 / 21.0 / 66.4  & 32.6 / 12.8 / 26.4 / 67.9 \\ \hline
\multirow{3}{*}{Zh}   & mT5 (580M)                & 30.9 / 9.8 / 23.1 / 66.5  & \cellcolor{ggray!90}{31.1 / 13.1 / 24.9 / 66.5} & 31.2 / 9.2 / 24.7 / 68.3  & 32.5 / 11.6 / 26.8 / 67.0 & \cellcolor{ggray!90}{29.2 / 9.3 / 23.8 / 62.5}  & 24.0 / 9.2 / 21.7 / 66.8  & 29.8 / 10.4 / 24.2 / 66.3 \\
                      & mBART (610M)              & 32.4 / 10.6 / 24.4 / 67.4 & \cellcolor{ggray!90}{35.7 / 17.7 / 28.6 / 68.5} & 33.5 / 10.9 / 27.5 / 69.5 & 33.1 / 11.6 / 26.9 / 67.2 & \cellcolor{ggray!90}{35.3 / 15.2 / 28.6 / 64.5} & 25.3 / 9.3 / 22.2 / 67.3  & 32.6 / 12.5 / 26.4 / 67.4 \\
                      & Pisces (610M)             & 33.4 / 11.0 / 25.5 / 68.2 & \cellcolor{ggray!90}{39.1 / 22.5 / 32.5 / 70.7} & 34.6 / 11.4 / 27.9 / 70.2 & 34.2 / 12.3 / 27.8 / 67.6 & \cellcolor{ggray!90}{37.7 / 17.4 / 31.1 / 65.6} & 27.8 / 11.1 / 24.3 / 68.0 & 34.5 / 14.3 / 28.2 / 68.4 \\ \hline
\multirow{3}{*}{Th}   & mT5 (580M)                & 25.7 / 6.6 / 19.1 / 62.6  & \cellcolor{ggray!90}{25.3 / 9.9 / 19.8 / 63.5}  & \cellcolor{ggray!90}{30.3 / 10.3 / 24.6 / 66.4} & \cellcolor{ggray!90}{27.2 / 7.6 / 22.8 / 64.1}  & 33.7 / 13.2 / 25.6 / 63.4 & \cellcolor{ggray!90}{22.0 / 9.5 / 19.7 / 63.9}  & 27.4 / 9.5 / 21.9 / 64.0  \\
                      & mBART (610M)              & 27.1 / 6.9 / 19.7 / 63.7  & \cellcolor{ggray!90}{27.6 / 11.0 / 21.5 / 64.2} & \cellcolor{ggray!90}{30.8 / 12.5 / 25.1 / 66.6} & \cellcolor{ggray!90}{33.1 / 15.8 / 28.4 / 67.5} & 35.9 / 14.7 / 26.9 / 65.2 & \cellcolor{ggray!90}{27.2 / 13.0 / 24.2 / 66.4} & 30.3 / 12.3 / 24.3 / 65.6 \\
                      & Pisces (610M)             & 29.5 / 9.1 / 21.7 / 65.6  & \cellcolor{ggray!90}{38.0 / 19.6 / 30.4 / 69.4} & \cellcolor{ggray!90}{35.8 / 15.6 / 29.1 / 69.2} & \cellcolor{ggray!90}{37.1 / 18.5 / 32.3 / 69.2} & 36.4 / 15.2 / 27.3 / 65.5 & \cellcolor{ggray!90}{30.2 / 15.4 / 26.7 / 68.2} & 34.5 / 15.6 / 27.9 / 67.8 \\ \hline
\multirow{3}{*}{Tr}   & mT5 (580M)                & 29.4 / 10.1 / 22.4 / 67.1 & \cellcolor{ggray!90}{32.9 / 12.2 / 24.6 / 67.2} & 29.2 / 7.8 / 23.8 / 67.1  & 30.1 / 9.5 / 25.1 / 67.6  & \cellcolor{ggray!90}{30.7 / 11.3 / 24.9 / 62.8} & 28.0 / 11.9 / 24.2 / 67.2 & 30.0 / 10.5 / 24.2 / 66.5 \\
                      & mBART (610M)              & 32.0 / 11.1 / 24.9 / 68.0 & \cellcolor{ggray!90}{36.0 / 17.2 / 28.9 / 68.9} & 32.5 / 9.5 / 26.0 / 68.6  & 31.5 / 9.8 / 25.5 / 67.9  & \cellcolor{ggray!90}{37.5 / 18.1 / 31.1 / 66.4} & 28.8 / 12.7 / 24.9 / 67.8 & 33.1 / 13.1 / 26.9 / 67.9 \\
                      & Pisces (610M)             & 33.3 / 11.5 / 25.3 / 68.6 & \cellcolor{ggray!90}{38.3 / 18.9 / 30.8 / 70.2} & 33.2 / 10.2 / 26.6 / 69.2 & 32.2 / 10.1 / 26.0 / 68.7 & \cellcolor{ggray!90}{40.9 / 22.0 / 34.3 / 67.7} & 30.8 / 14.0 / 26.5 / 68.5 & 34.8 / 14.4 / 28.2 / 68.8 \\ \hline
\multirow{3}{*}{Avg.} & mT5 (580M)                & 29.4 / 9.0 / 22.2 / 66.0  & 30.1 / 11.4 / 23.2 / 66.0 & 31.0 / 9.9 / 25.2 / 67.5  & 30.6 / 10.9 / 25.7 / 66.8 & 31.1 / 11.5 / 24.8 / 62.7 & 24.5 / 9.8 / 21.7 / 66.0  & 29.4 / 10.4 / 23.8 / 65.8 \\
                      & mBART (610M)              & 31.0 / 9.6 / 23.3 / 66.8  & 32.8 / 14.1 / 25.6 / 67.1 & 33.1 / 11.8 / 26.7 / 68.4 & 33.0 / 13.0 / 27.7 / 67.9 & 35.1 / 15.2 / 28.1 / 64.9 & 26.6 / 11.2 / 23.3 / 67.1 & 31.9 / 12.5 / 25.8 / 67.0 \\
                      & Pisces (610M)             & 32.2 / 10.4 / 24.3 / 67.7 & 36.2 / 17.3 / 28.7 / 69.0 & 35.1 / 13.3 / 28.4 / 69.5 & 35.1 / 14.3 / 29.5 / 69.1 & 37.9 / 18.4 / 30.7 / 66.5 & 29.0 / 13.2 / 25.4 / 68.1 & 34.2 / 14.5 / 27.8 / 68.3 \\ \toprule[1pt]                        
\end{tabular}
}

\caption{Experimental results on CrossSum (\textsc{Rouge-1} / \textsc{Rouge-2} / \textsc{Rouge-l} / \textsc{BertScore}). \colorbox{ggray!90}{gray} indicates the \colorbox{ggray!90}{zero-shot} directions. ``Avg.'' denotes the average scores w.r.t each row, each column or all directions.}
\label{table:res_crossum}
\vspace{-0.3cm}
\end{table*}

\noindent \textbf{Fine-Tuning and Testing Details.} In the fine-tuning stage, we fine-tune the \textsc{Pisces} model on 8 NVIDIA Tesla V100 GPUs (32G) with 4 batch size, 10 epochs, 2K warmup steps, 3e-5 learning rate, and set the maximum number of tokens for input sequences to 1024.
To balance the high-resource and low-resource language data, following~\citet{xue-etal-2021-mt5}, we sample the training examples according to the probability $p(D)\propto|D|^{\alpha}$, where $p(D)$ is the probability of sampling training examples from a give direction during fine-tuning and $|D|$ is the number of original examples in the direction. We set the hyperparameter $\alpha$ to 0.5.
To fine-tune mT5 baseline on M2MS, the language tags (\emph{e.g.}, \texttt{<En>} and \texttt{<Zh>}) are appended at the inputs of both encoder and decoder sides.
In the test process, we set the beam size and the maximum decoded length to 5 and 128, respectively.

\section{Experiments on CrossSum}

\subsection{Data Statistics.}
\label{appendix:crosssum1}
Table~\ref{table:crosssum_statistics} lists the data statistics of the CrossSum dataset~\cite{Hasan2021CrossSumBE} used in our experiments. The data splitting mainly inherits from the original CrossSum except for zero-shot directions and monolingual directions: (1) If the number of samples in a direction (\emph{e.g.}, Fr$\Rightarrow$Hi) is less than 1k, we will regard the direction as a zero-shot direction and evenly split its samples into validation and test sets. (2) Considering the number of samples in cross-lingual directions is hundred-level or thousand-level, we truncate the number of samples in each monolingual direction (\emph{e.g.}, En$\Rightarrow$En) to 10k to make a balance. The corresponding splitting follows 8:1:1. If the number of samples in a monolingual direction (\emph{e.g.}, Th$\Rightarrow$Th) is less than 10k, its splitting follows the original CrossSum.

\begin{table*}[t]
\centering
\resizebox{0.95\textwidth}{!}
{
\begin{tabular}{c|l|c|c|c|c|c|c}
\toprule[1pt]
\multicolumn{1}{l|}{\diagbox[dir=NW]{Src}{Trg}} & \multicolumn{1}{c|}{Model}       & En                        & Fr                        & Hi                        & Zh                        & Th                        & Tr                     \\ \bottomrule[1pt]
\multirow{3}{*}{En}    & mT5                            & \cellcolor{ggreen!90}{40.9 / 17.7 / 34.2} & \cellcolor{ggreen!90}{36.0 / 15.0 / 29.4} & \cellcolor{ggreen!30}{30.1 / 8.9 / 23.5}  & \cellcolor{ggreen!90}{37.0 / 13.4 / 32.1} & \cellcolor{ggreen!30}{38.6 / 17.8 / 33.4} & \cellcolor{ggray!90}{3.3 / 0.2 / 3.0} \\
  & mBART                            & \cellcolor{ggreen!90}{41.9 / 18.2 / 34.9} & \cellcolor{ggreen!90}{37.2 / 15.8 / 30.3} & \cellcolor{ggreen!30}{31.7 / 9.6 / 24.5}  & \cellcolor{ggreen!90}{37.9 / 13.9 / 32.7} & \cellcolor{ggreen!30}{39.5 / 18.5 / 34.0} & \cellcolor{ggray!90}{3.2 / 0.2 / 3.0} \\
& \textsc{Pisces} & \cellcolor{ggreen!90}{42.8 / 18.8 / 35.5} & \cellcolor{ggreen!90}{38.1 / 16.4 / 31.1} & \cellcolor{ggreen!30}{33.7 / 10.8 / 26.6} & \cellcolor{ggreen!90}{38.8 / 14.2 / 33.3} & \cellcolor{ggreen!30}{40.9 / 19.3 / 35.6} & \cellcolor{ggray!90}{4.5 / 0.7 / 4.2} \\ \hline
\multirow{3}{*}{Fr}    & mT5                            & \cellcolor{ggreen!90}{37.0 / 14.3 / 31.2} & \cellcolor{ggreen!90}{38.6 / 18.3 / 32.6} & \cellcolor{ggray!90}{27.0 / 7.4 / 22.1}  & \cellcolor{ggreen!90}{35.6 / 12.4 / 31.3} & \cellcolor{ggreen!30}{36.4 / 15.9 / 32.1} & \cellcolor{ggray!90}{3.1 / 0.2 / 2.9} \\
 & mBART                            & \cellcolor{ggreen!90}{38.2 / 15.0 / 31.7} & \cellcolor{ggreen!90}{39.2 / 17.9 / 32.0} & \cellcolor{ggray!90}{28.7 / 7.9 / 22.3}  & \cellcolor{ggreen!90}{36.9 / 12.8 / 31.6} & \cellcolor{ggreen!30}{37.9 / 16.6 / 32.6} & \cellcolor{ggray!90}{3.1 / 0.2 / 3.0} \\
                       & \textsc{Pisces} & \cellcolor{ggreen!90}{39.2 / 15.4 / 32.4} & \cellcolor{ggreen!90}{40.0 / 18.3 / 32.5} & \cellcolor{ggray!90}{31.3 / 8.8 / 24.2}  & \cellcolor{ggreen!90}{37.4 / 13.0 / 31.9} & \cellcolor{ggreen!30}{39.2 / 17.3 / 33.6} & \cellcolor{ggray!90}{4.1 / 0.6 / 3.8} \\ \hline
\multirow{3}{*}{Hi}    & mT5                            & \cellcolor{ggreen!30}{36.9 / 14.2 / 30.3} & \cellcolor{ggray!90}{31.9 / 11.6 / 25.6} & \cellcolor{ggreen!30}{34.9 / 11.9 / 27.2} & \cellcolor{ggray!90}{32.1 / 9.6 / 27.5} & \cellcolor{ggray!90}{34.0 / 14.1 / 29.2} & \cellcolor{ggray!90}{3.2 / 0.3 / 3.0} \\
 & mBART                            & \cellcolor{ggreen!30}{37.9 / 14.6 / 30.8} & \cellcolor{ggray!90}{32.8 / 12.2 / 25.9} & \cellcolor{ggreen!30}{35.6 / 12.5 / 27.8} & \cellcolor{ggray!90}{33.2 / 10.6 / 28.2} & \cellcolor{ggray!90}{35.4 / 14.6 / 30.1} & \cellcolor{ggray!90}{3.4 / 0.3 / 3.2} \\
& \textsc{Pisces} & \cellcolor{ggreen!30}{39.8 / 16.0 / 32.7} & \cellcolor{ggray!90}{35.7 / 14.1 / 28.4} & \cellcolor{ggreen!30}{37.2 / 13.6 / 28.8} & \cellcolor{ggray!90}{35.9 / 11.8 / 30.7} & \cellcolor{ggray!90}{38.1 / 16.6 / 32.6} & \cellcolor{ggray!90}{4.0 / 0.6 / 3.8} \\ \hline
\multirow{3}{*}{Zh}    & mT5                            & \cellcolor{ggreen!90}{38.3 / 14.3 / 31.5} & \cellcolor{ggreen!90}{34.5 / 13.8 / 28.3} & \cellcolor{ggray!90}{26.0 / 6.2 / 20.3}  & \cellcolor{ggreen!90}{41.1 / 16.5 / 35.5} & \cellcolor{ggray!90}{36.1 / 14.7 / 30.8} & \cellcolor{ggray!90}{3.3 / 0.3 / 3.2} \\
  & mBART                            & \cellcolor{ggreen!90}{39.2 / 15.1 / 32.0} & \cellcolor{ggreen!90}{36.0 / 14.5 / 29.0} & \cellcolor{ggray!90}{27.0 / 6.6 / 20.8}  & \cellcolor{ggreen!90}{41.7 / 17.0 / 35.9} & \cellcolor{ggray!90}{36.8 / 15.3 / 31.4} & \cellcolor{ggray!90}{3.4 / 0.2 / 3.2} \\
& \textsc{Pisces} & \cellcolor{ggreen!90}{40.3 / 15.8 / 33.0} & \cellcolor{ggreen!90}{37.4 / 15.4 / 29.9} & \cellcolor{ggray!90}{29.6 / 8.2 / 23.2}  & \cellcolor{ggreen!90}{42.5 / 17.5 / 36.3} & \cellcolor{ggray!90}{39.2 / 17.0 / 33.6} & \cellcolor{ggray!90}{4.3 / 0.6 / 4.0} \\ \hline
\multirow{3}{*}{Th}    & mT5                            & \cellcolor{ggreen!30}{37.6 / 15.0 / 30.7} & \cellcolor{ggreen!30}{34.1 / 13.9 / 27.8} & \cellcolor{ggray!90}{26.2 / 6.8 / 20.5}  & \cellcolor{ggray!90}{34.0 / 11.0 / 28.7} & \cellcolor{ggreen!30}{41.1 / 20.3 / 36.0} & \cellcolor{ggray!90}{3.3 / 0.3 / 3.2} \\
 & mBART                            & \cellcolor{ggreen!30}{38.5 / 15.4 / 31.9} & \cellcolor{ggreen!30}{35.6 / 14.2 / 28.3} & \cellcolor{ggray!90}{27.8 / 7.3 / 21.4}  & \cellcolor{ggray!90}{34.6 / 11.3 / 29.0} & \cellcolor{ggreen!30}{42.2 / 20.8 / 36.2} & \cellcolor{ggray!90}{3.3 / 0.3 / 3.1} \\
& \textsc{Pisces} & \cellcolor{ggreen!30}{40.2 / 16.6 / 33.2} & \cellcolor{ggreen!30}{37.2 / 15.4 / 29.7} & \cellcolor{ggray!90}{31.0 / 9.3 / 23.9}  & \cellcolor{ggray!90}{36.9 / 12.7 / 31.3} & \cellcolor{ggreen!30}{43.3 / 21.7 / 37.5} & \cellcolor{ggray!90}{4.3 / 0.7 / 4.0} \\ \hline
\multirow{3}{*}{Tr}    & mT5                            & \cellcolor{ggray!90}{14.2 / 2.2 / 12.1}  & \cellcolor{ggray!90}{14.9 / 2.9 / 12.2}  & \cellcolor{ggray!90}{11.2 / 1.4 / 10.8}  & \cellcolor{ggray!90}{19.2 / 2.6 / 15.9}  & \cellcolor{ggray!90}{20.0 / 4.2 / 17.9} & \cellcolor{ggray!90}{3.0 / 0.2 / 2.8} \\
 & mBART                            & \cellcolor{ggray!90}{15.7 / 2.6 / 13.4}  & \cellcolor{ggray!90}{16.0 / 3.2 / 13.2}  & \cellcolor{ggray!90}{14.9 / 2.3 / 12.6}  & \cellcolor{ggray!90}{19.9 / 3.0 / 17.6}  & \cellcolor{ggray!90}{21.4 / 4.8 /  19.3} & \cellcolor{ggray!90}{3.1 / 0.2 / 3.0} \\
& \textsc{Pisces} & \cellcolor{ggray!90}{28.3 / 8.8 / 23.4}  & \cellcolor{ggray!90}{27.3 / 9.3 / 22.2}  & \cellcolor{ggray!90}{23.2 / 5.5 / 18.5}  & \cellcolor{ggray!90}{29.8 / 8.2 / 25.7}  & \cellcolor{ggray!90}{30.8 / 11.3 / 26.7} & \cellcolor{ggray!90}{5.3 / 0.8 / 5.0} \\ \toprule[1pt]
\end{tabular}
}

\caption{Experimental results on WikiLingua (\textsc{Rouge-1} / \textsc{Rouge-2} / \textsc{Rouge-l}). \colorbox{ggreen!90}{Green}, \colorbox{ggreen!30}{light green} and \colorbox{ggray!90}{gray} indicate the \colorbox{ggreen!90}{high-resource}, \colorbox{ggreen!30}{low-resource} and \colorbox{ggray!90}{zero-shot} directions, respectively.} 
\label{table:full_res}
\vspace{-0.3cm}
\end{table*}

\subsection{Experimental Results.}
\label{appendix:crosssum2}
Table~\ref{table:res_crossum} shows the experimental results on CrossSum. Our \textsc{Pisces} outperforms mBART-50 by 2.3 \textsc{Rouge-1}, 2.0 \textsc{Rouge-2}, 2.0 \textsc{Rouge-l} and 1.3 \textsc{BertScore} in the average of all directions, which verifies the effectiveness of \textsc{Pisces}. For the average results in all zero-shot directions, mBART-50 achieves 33.8, 15.7, 28.1 and 67.1 in terms of \textsc{Rouge-1/2/l} and \textsc{BertScore}. The counterparts of \textsc{Pisces} are 37.9, 19.6, 31.8 and 69.3, showing its superiority in the zero-shot directions.

\section{Full Results on WikiLingua}
\label{appendix:rouge}
Table~\ref{table:full_res} shows the experimental results in terms of \textsc{Rouge-1}, \textsc{Rouge-2} and \textsc{Rouge-l}, respectively.

\section{Ablations in Conventional Zero-Shot Directions}
\label{appendix:ablations}

\begin{table}[t]
\centering
\resizebox{0.48\textwidth}{!}
{
\begin{tabular}{lcccc}
\bottomrule[1pt]
             & Fr$\Rightarrow$Hi       & Hi$\Rightarrow$Fr       & Hi$\Rightarrow$Zh       & Zh$\Rightarrow$Hi       \\ \toprule[1pt]
\textsc{Pisces}       & \textbf{21.4} / \textbf{69.1} & \textbf{26.1} / \textbf{72.9} & \textbf{26.1} / \textbf{70.4} & \textbf{20.3} / \textbf{68.5} \\
\quad w/o TS & 20.7 / 68.6 & 25.2 / 72.8 & 25.1 / 69.9 & 19.5 / 67.9 \\
\quad w/o CL & 20.6 / 68.8 & 25.2 / \textbf{72.9} & 25.3 / 70.0 & 19.5 / 67.8 \\
\quad w/o TS \& CL   & 19.6 / 68.1 & 23.6 / 72.1 & 24.0 / 69.1 & 18.1 / 66.9 \\ \bottomrule[1pt]
\end{tabular}
}

\resizebox{0.48\textwidth}{!}
{
\begin{tabular}{lcccc}
\bottomrule[1pt]
             & Hi$\Rightarrow$Th       & Th$\Rightarrow$Hi       & Zh$\Rightarrow$Th       & Th$\Rightarrow$Zh       \\ \toprule[1pt]
\textsc{Pisces}     & \textbf{29.1} / \textbf{68.5} & \textbf{21.4} / \textbf{69.0} & \textbf{29.9} / \textbf{68.9} & \textbf{27.0} / \textbf{71.0} \\
\quad w/o TS & 28.2 / 68.1 & 20.3 / 68.3 & 28.7 / 68.3 & 25.8 / 70.3 \\
\quad w/o CL & 28.0 / 68.0 & 20.3 / 68.4 & 29.0 / 68.5 & 26.0 / 70.4 \\
\quad w/o TS \& CL      & 26.7 / 67.4 & 18.8 / 67.4 & 27.8 / 67.6  & 25.0 / 69.4 \\ \bottomrule[1pt]
\end{tabular}
}

\caption{Results of ablation studies.} 
\label{table:ablations_full}
\vspace{-0.3cm}
\end{table}

Table~\ref{table:ablations_full} shows the ablation results in all conventional zero-shot directions.

\section{Error Analysis}
\label{appendix:errors}

We first randomly select 100 summaries generated by \textsc{Pisces} on WikiLingua (En$\Rightarrow$Zh). After manually examining the generated summaries, we find the following major error types:
\begin{itemize}[leftmargin=*,topsep=0pt]
\setlength{\itemsep}{0pt}
\setlength{\parsep}{0pt}
\setlength{\parskip}{0pt}
\item \textbf{Missing Information}: part of the information in the ground truth summary is not mentioned in the generated summary. This is the most frequent error type, and accounts for 39\% of the generated summaries.
\item \textbf{Faithfulness}: the generated summary involves information that is inconsistent with (or not presented in) the source document. We find 32\% of the summaries have this error.
\item \textbf{Redundancy}: the generated summary contains additional information beyond the ground truth summary. 17\% of the generated summaries contain this error.
\item \textbf{Foreign Words}: the generated summary involves words in another language. 9\% of the generated Chinese summaries involve some (typically one or two) words in another language.
\end{itemize}

Redundancy and missing information are two major flaws caused by the limited summarization ability~\cite{johner-etal-2021-error}. Faithfulness error is another error type that has been noticed in the summarization research field recently~\cite{huang2021factual}. The neural generative summarization models are highly prone to generate factual inconsistency errors~\cite{huang2021factual}. Some studies~\cite{kryscinski-etal-2020-evaluating,wang2022analyzing} show that over 30\% of the summaries generated by neural models contain this error. We confirm that CLS also involves the faithfulness error. Future work could give deeper and more fine-grained analyses of this error type.

The issue of foreign words could also refer to the code-switching phenomenon~\cite{Pfaff1979ConstraintsOL}. Note that the generated foreign words are not limited in the source language. In several cases, the generated Chinese summaries of the given English documents even involve Thai words. We also find the semantics of these foreign words are typically coherent with their context. This error type might be caused by the cross-lingual pre-training (which bridges the representation gap of parallel words in different languages) in \textsc{Pisces}.

\end{document}